\documentclass{article} 
\usepackage{iclr2027_conference,times}

\usepackage{amsmath,amsfonts,bm}

\def\eqref#1{equation~\ref{#1}}

\def\1{\bm{1}}

\DeclareMathAlphabet{\mathsfit}{\encodingdefault}{\sfdefault}{m}{sl}
\SetMathAlphabet{\mathsfit}{bold}{\encodingdefault}{\sfdefault}{bx}{n}

\DeclareMathOperator*{\argmax}{arg\,max}

\usepackage{amsmath,amssymb,booktabs,graphicx}
\newcommand{\thr}{\tau}                 
\newcommand{\base}{\thr_{0}}            
\newcommand{\Kcl}{K_{\mathrm{reg}}}     
\newcommand{\clust}{c}                  
\newcommand{\lvl}{\ell}                 
\newcommand{\ladder}{\mathcal{M}}       
\newcommand{\save}{s}                   
\newcommand{\risk}{\rho}                
\newcommand{\Npts}{N}                   
\newcommand{\pairs}{P}                  

\usepackage{hyperref}
\usepackage{url}
\usepackage[capitalise]{cleveref}  

\title{TileSkipper: Region-Adaptive Tile Pruning for 3D Gaussian Splatting}

\author{
Jingxing Li, Yongjae Lee, Deliang Fan \\
School of Electrical, Computer and Energy Engineering \\
Arizona State University \\
Tempe, AZ, USA \\
\texttt{\{jingxing,ylee298,dfan\}@asu.edu}
\AND
Abhay Kumar Yadav, Cheng Peng, Rama Chellappa \\
Department of Electrical and Computer Engineering \\
Johns Hopkins University \\
Baltimore, MD, USA \\
\texttt{ayadav13@jh.edu, \{cpeng26,rchella4\}@jhu.edu}
}

\iclrfinalcopy
\date{}
\hypersetup{hidelinks,pdftitle={TileSkipper: Region-Adaptive Tile Pruning for 3D Gaussian Splatting}}

\begin{document}

        \maketitle
        \pagestyle{plain}
        \thispagestyle{plain}

        \begin{abstract}
Tiled 3D Gaussian Splatting rasterizers often use one scene-wide contribution
cutoff for tile enumeration, although content differs in its sensitivity to
support truncation. TileSkipper selects a static per-Gaussian cutoff policy for
a frozen checkpoint. Calibration renders measure candidate pair savings and
an isolated-removal distortion proxy that accounts for front transmittance and
background color. The method allocates cutoffs across 64 Gaussian groups and
accepts policies only after complete renders on disjoint selection views.
The exported policy uses one byte per Gaussian, with no parameter updates,
additional kernel, or per-frame policy inference. Across 13 scenes from
Mip-NeRF~360, Tanks~\&~Temples, and Deep Blending, a fixed-policy AccuTile
sweep gives dataset-macro speedups of $\mathbf{1.088\times}$ at standard
resolution and $\mathbf{1.238\times}$ at 3840 pixels wide, with
$-0.007/-0.023$\,dB mean PSNR change. Six integrations with existing
opacity-aware bounds yield $1.009\times$--$1.121\times$ compiler-only speedups.
For four ports from $3\sigma$ rasterizers, we separately attribute the prior
exact-bound transition and our incremental gain. Matched-quality ablations
show modest gains over scene-global calibration and parity with per-Gaussian
control; the standalone comparison with AdaGScale is regime-dependent.
\end{abstract}

        \section{Introduction}

        3D Gaussian Splatting (3DGS)~\citep{kerbl20233dgs} has become a dominant representation for real-time novel view synthesis. While fast in absolute terms, this pipeline pays a per-frame cost that scales with the number of \emph{(Gaussian, tile) pairs}: a Gaussian with a large projected footprint is replicated across many tiles, and the binning, sorting, and rasterization kernels must process every resulting pair. Reducing this work matters particularly for stereo rendering and for mobile or XR hardware with limited throughput.

                A natural way to shrink the work per frame is to test tile intersection against a tighter footprint than the one used for blending. AdR-Gaussian~\citep{wang2024adrgaussian}, Speedy-Splat~\citep{hanson2025speedysplat}, and FastGS~\citep{ren2025fastgs} use an opacity level set to replace a conservative projected footprint by a tighter bound. We borrow this level-set construction; the formula itself is not our contribution. Our question begins after the bound is available: how far can the cutoff be raised, differently for each frozen Gaussian, under a measured image-error budget? Raising $\thr$ removes low-contribution tails, but their cumulative image error is not guaranteed to be negligible and must be measured. Several threshold-based paths expose only a \emph{single global} $\thr$; whether fixed by the implementation or tuned for a dataset, it cannot allocate an error budget among Gaussians. Flat backgrounds, distant geometry, and
                nearly opaque material patches tolerate aggressive tile skipping;
                sharp edges, fine textures, and specular highlights require a
                conservative threshold. A global rule is forced into a compromise
                that is too loose on the easy regions and too tight on the hard ones.
                We make this concrete in
                \Cref{app:local_tolerance}: on a single \emph{playroom} view, the
                largest locally safe threshold spans the full tested range from
                $\thr^\star=1/255$ to $16/255$ across regions of the same image.

                AdaGScale~\citep{jo2026adagscale} is the closest prior method to this
                operation. It also performs lossy, viewpoint-adaptive scaling only for
                tile intersection, uses offline parameter search and a depth lookup table,
                and retains the original Gaussian during color accumulation. Consequently,
                training-free operation, a lookup table, and sub-Gaussian support control
                are not distinctive claims of TileSkipper. The difference is the decision
                principle: AdaGScale derives a current-view peripheral score from local
                Gaussian contribution and depth, whereas TileSkipper measures candidate
                deletion residuals over calibration views using the transmittance in front
                of the Gaussian and the composited color behind it, allocates a discrete
                error budget across Gaussian groups, and rejects proposed policies by complete renders on
                disjoint training views. TileSkipper trades AdaGScale's runtime view
                adaptation for a scene-level, static one-byte policy with no per-frame
                policy inference. The policy is not indexed by precompiled camera poses:
                an arbitrary runtime camera is projected and rendered normally using the
                same level byte, so an interactive viewer requires no camera matching.
                The checkpoint-matched comparison in \Cref{tab:adagscale_official}
                is dataset-dependent and does not establish broad runtime dominance.

        \begin{figure}[t]
            \centering
            \includegraphics[width=\linewidth]{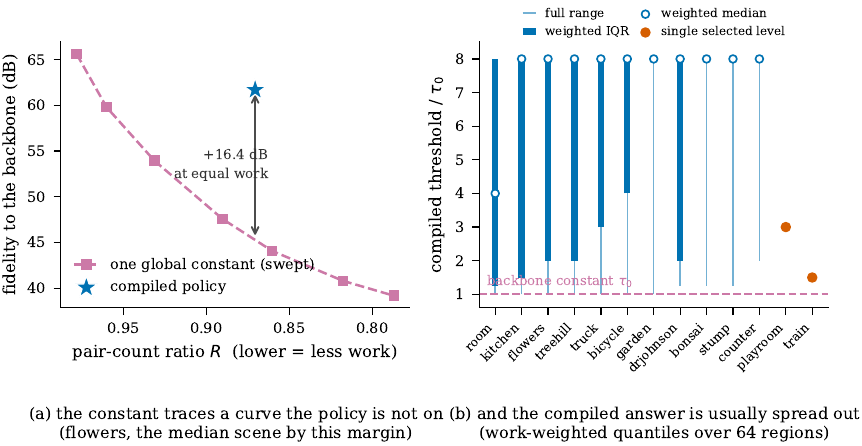}
            \caption{\textbf{A scene-global cutoff leaves allocation slack.}
            \emph{(a)} On this diagnostic work--fidelity frontier, the compiled regional policy
            reaches $16.4$\,dB higher teacher PSNR than the swept scene-global cutoff at the same
            pair count; this is the median scene by that margin. The number measures approximation
            fidelity to the same backbone under equal enumerated work, not a $16.4$\,dB improvement
            in ground-truth PSNR. At matched teacher PSNR, Regional is
            $1.040$--$1.046\times$ faster than \textsc{Ours-Global} at standard resolution and
            $1.041$--$1.083\times$ faster at 4K (\Cref{tab:matched}); it ties the per-Gaussian
            variant. \emph{(b)} Work-weighted quartiles (thick lines and dots) and
            full selected ranges (thin lines) over $64$ regions. Eleven of thirteen scenes
            select more than one level; seven span the full $\thr_0$--$8\thr_0$ range, and
            two collapse to one level. This dispersion shows what a scene-global constant
            cannot represent; it does not by itself measure end-to-end speedup.}
            \label{fig:motivation}
        \end{figure}

                Figure~\ref{fig:motivation} motivates heterogeneous allocation.
                Calibration renders measure pair savings and a compositing-aware
                isolated-removal residual for each cutoff. \textsc{Ours-Global}
                selects one scene-wide value; \textsc{Ours-Regional} allocates levels
                across risk-stratified spatial groups. Complete renders on disjoint
                training views validate every proposal. Regional allocation adds
                roughly $4$--$5\%$ over Global at matched native-resolution quality
                (\Cref{tab:matched}). The output is one \texttt{uint8} level per
                Gaussian plus a lookup table, with no checkpoint update or per-frame
                policy inference. \Cref{tab:resolution_sweep} evaluates the same
                exported policies across resolutions, while \Cref{ssec:exp_global}
                compares against fixed-threshold alternatives.

                The method is deliberately a post-training systems adapter. Its measurements are scene-specific, but the resulting decision is static and portable: policies compiled on AccuTile transfer unchanged into SeeLe~\citep{huang2026seele} and FlashGS~\citep{feng2025flashgs}. A measured cost model further separates the pair-dependent work the policy can reduce from fixed work it cannot, and predicts when a complementary geometric pre-culler pays.

\paragraph{Contributions.}
\begin{itemize}
\item \textbf{Compositing-aware support allocation for frozen checkpoints.} For each discrete cutoff, TileSkipper measures exact pair savings and an isolated-removal residual that accounts for front transmittance and the composited color behind the Gaussian, then allocates the image-error budget across risk-stratified, workload-balanced regions. Complete renders on disjoint training views reject proxy failures. The technical distinction from AdaGScale is this measured compositing residual and the resulting error allocation; static one-byte export with no per-frame policy inference is a deployment tradeoff, not the claimed novelty.
\item \textbf{Evidence across model and rasterizer families.} Six integrations isolate the compiler on paths that already expose an opacity-aware tile bound. Four additional ports begin from released $3\sigma$ rasterizers; for these rows we report the deployment-facing total and separately attribute the prior exact-bound transition and the incremental TileSkipper gain. They use the same nine Mip-NeRF~360~\citep{barron2022mipnerf360}, two Tanks~\&~Temples~\citep{knapitsch2017tnt}, and two Deep Blending~\citep{hedman2018db} scenes. A standalone comparison with AdaGScale fixes the renderer, checkpoint-matched calibration table, selection views, test views, and timing protocol. A Speedy-Splat sweep varies $\Kcl$, the number of policy regions, from 1 to 1024 to separate scene calibration from regional allocation.
\item \textbf{A measured boundary on systems benefit.} A latency model fitted from controlled sweeps predicts the pair-dependent ceiling and the complementary pre-culling regime. The optional pre-culler is image-bit-exact in all 52 measured runs, but is correctly disabled when its planning overhead exceeds the predicted saving.
\end{itemize}

\section{Related Work}
                        \label{sec:related}

                        We distinguish primitive, tile-pair, and fragment reduction; Appendix~\ref{app:related_table} summarizes their control units.

                        \paragraph{Per-Gaussian thresholds at the fragment level.} Fragment Pruning~\citep{ye2024fragmentpruning} learns one truncation threshold per Gaussian via a sigmoid relaxation and controls per-pixel coverage after sorting. TileSkipper selects discrete tile-enumeration cutoffs without updating checkpoint parameters and acts before sorting, removing pairs from both sorting and blending. \Cref{app:fragment_backbones} evaluates TileSkipper on the checkpoint families used by Fragment Pruning; it tests checkpoint compatibility, not composition with learned fragment pruning.

                        \paragraph{System-level paths.} SeeLe~\citep{huang2026seele} combines camera-cluster preprocessing with contribution-aware rasterization. Its tile-intersection test uses the support in \Cref{eq:support_radius} with $\thr=1/255$ fixed at compile time. TileSkipper changes that cutoff policy, while our transfer experiment retains SeeLe's other components and does not claim to reproduce its full mobile deployment pipeline.

                        \paragraph{Tile-pair reduction.} AdR-Gaussian~\citep{wang2024adrgaussian} derives the opacity level set at the renderer's minimum splatting-opacity cutoff and bounds it with an adaptive radius and an axis-aligned box. We borrow that level-set construction rather than claim the support formula as new. Speedy-Splat~\citep{hanson2025speedysplat} introduces accurate tile pruning, FastGS~\citep{ren2025fastgs} combines it with a compact box, and FlashGS~\citep{feng2025flashgs} and QuadBox~\citep{li2026quadbox} further tighten geometric enumeration. These exact or fixed-threshold bounds are complementary to selecting heterogeneous lossy cutoffs.

                        \paragraph{View-adaptive tile support.} AdaGScale~\citep{jo2026adagscale} is the closest work in action space: it estimates a current-view peripheral contribution score, applies depth-dependent lookup-table calibration, and adjusts the scale used for tile intersection while retaining the original Gaussian for color accumulation. TileSkipper instead measures a compositing residual---including front transmittance and the composited color behind the Gaussian---for discrete candidate cutoffs over calibration views, allocates the resulting error budget across Gaussian groups, and accepts a static policy only after complete-image validation on disjoint training views. AdaGScale offers runtime view adaptation; TileSkipper incurs no per-frame policy inference and directly measures the compiled policy's complete-render error. \Cref{tab:adagscale_official} compares the standalone methods inside AdaGScale's released renderer.

\section{Method}
\label{sec:method}

\begin{figure}[t]
  \centering
  \includegraphics[width=\linewidth]{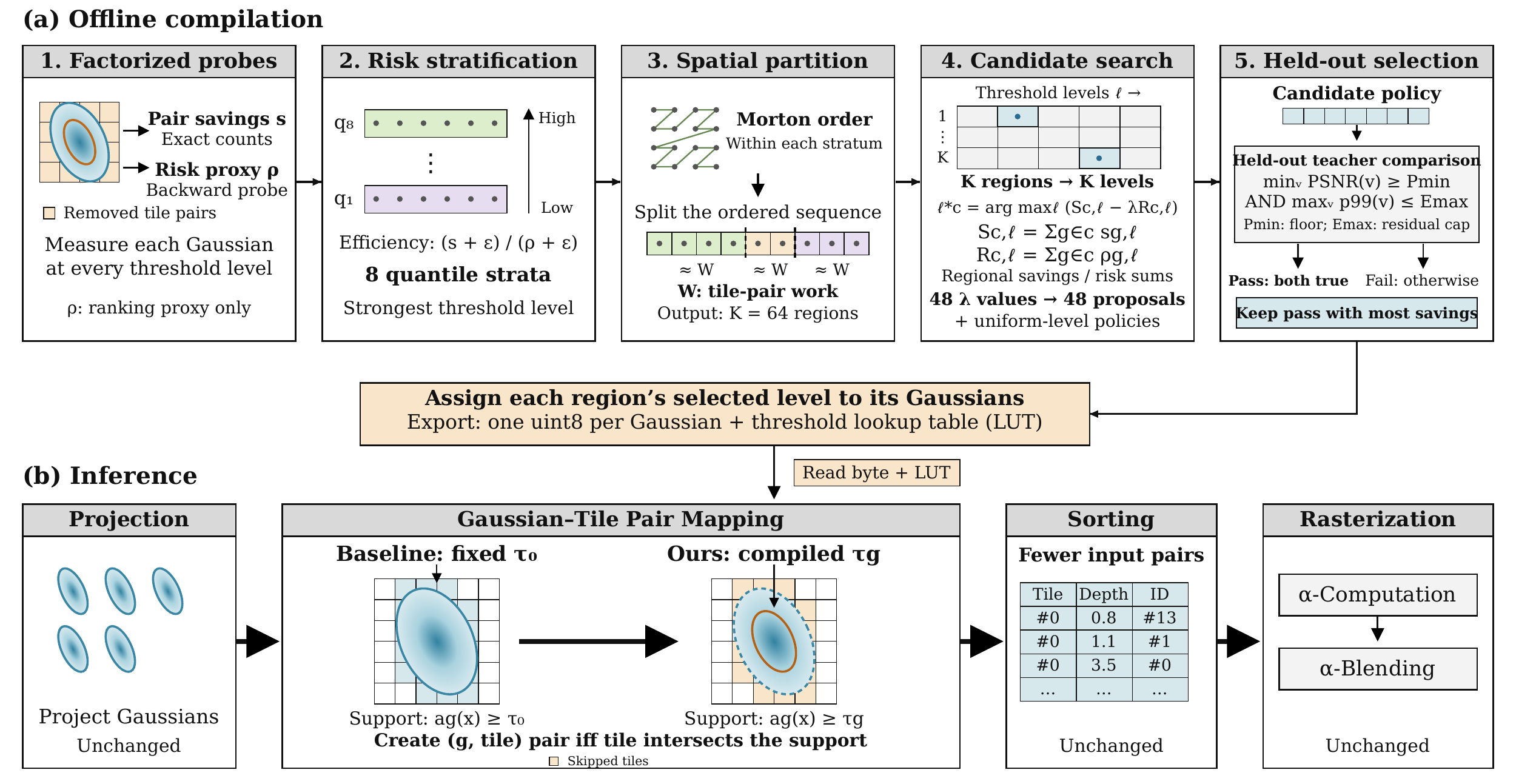}
  \caption{\textbf{Offline compilation and runtime application of TileSkipper.}
  The diagram abbreviates the region-count symbol $\Kcl$ as $K$.
  \textbf{(a)} Once per scene, the compiler uses a frozen 3DGS checkpoint and
  calibration views to measure per-Gaussian pair savings $s$ and isolated-removal
  risk proxies $\rho$ for each threshold level. The backward probe collects
  sensitivities without updating any model parameter. Gaussians are assigned to
  eight efficiency-quantile strata $q_1,\ldots,q_8$ (low to high). Within each
  stratum, Morton ordering and workload-balanced contiguous partitioning produce
  $\Kcl=64$ regions in total; Gaussian counts may differ between regions. For region
  $c$ and level $\ell$,
  $S_{c,\ell}=\sum_{g\in c}s_{g,\ell}$ and
  $R_{c,\ell}=\sum_{g\in c}\rho_{g,\ell}$ aggregate pair savings and risk. Each
  penalty $\lambda$ selects one level for each of the $\Kcl$ regions. The 48
  resulting proposals and uniform-level policies are rendered on disjoint
  held-out training views. A policy passes only if its minimum per-view teacher
  PSNR meets $P_{\min}$ and its worst-view p99 absolute residual does not exceed
  $E_{\max}$; among passing policies, the compiler selects the one with the
  greatest pair savings. Teacher comparisons use the unmodified backbone; p99 is
  computed from maximum absolute RGB-channel residuals on a fixed $4\times$
  spatial subsample. Test views are never used for selection.
  \textbf{(b)} Each Gaussian inherits its region's selected level, exported as one
  \texttt{uint8} per Gaussian plus a threshold lookup table. At inference,
  the default threshold is $\tau_g=m_g\tau_0$, where $m_g$ is the selected
  multiplier and $\tau_0$ is the baseline threshold. Backend-specific handling
  of opacity caps is described in \Cref{app:threshold_semantics}. The
  contribution field $a_g(x)$ defines the thresholded support
  $\{x:a_g(x)\geq\tau\}$. A Gaussian--tile pair is created when the tile intersects
  this support, using $\tau_0$ for the baseline and $\tau_g$ for TileSkipper. The
  dashed blue contour denotes the original support, the orange contour the
  compiled support, and shaded orange tiles the skipped pairs. Gaussian parameters
  remain unchanged. The byte and lookup table are read inside tile enumeration,
  with no per-frame policy inference or additional kernel; projection, sorting,
  and blending remain unchanged. Footprints, partitions, and table entries are
  schematic.}
  \label{fig:method}
\end{figure}

TileSkipper takes a converged 3DGS checkpoint and a tile rasterizer whose
tile-intersection test uses one scene-wide contribution cutoff $\base$, then
replaces that constant with a static per-Gaussian cutoff policy. No checkpoint
parameter is retrained. At runtime the rasterizer reads one byte per Gaussian
and a lookup table, with no camera-conditioned policy inference.
\Cref{sec:prelim} defines the cutoff and its geometric effect;
\Cref{sec:compiler} measures pair savings and compositing risk, groups Gaussians
into workload-balanced regions, selects one cutoff level per region, and
validates complete candidate renders. Appendix~\ref{app:systems_details}
provides the empirical cost model, optional pre-culler, and dispatch rule.
\Cref{fig:method} summarizes the offline measurements, policy selection, and
the one-byte runtime application.

\emph{Official-Fixed} uses the published cutoff $\base$ for every Gaussian.
\textsc{Ours-Global} selects one scene-specific multiplier shared by all
Gaussians ($\Kcl=1$), whereas \textsc{Ours-Regional} selects levels for
$\Kcl=64$ regions. Here $\Kcl$ is the number of independently selected policy
regions, and $m_g$ denotes Gaussian $g$'s multiplier on $\base$; $m_g=1$
recovers Official-Fixed.

\begin{figure}[t]
  \centering
  \makebox[\linewidth][c]{\includegraphics[width=1.0\linewidth]{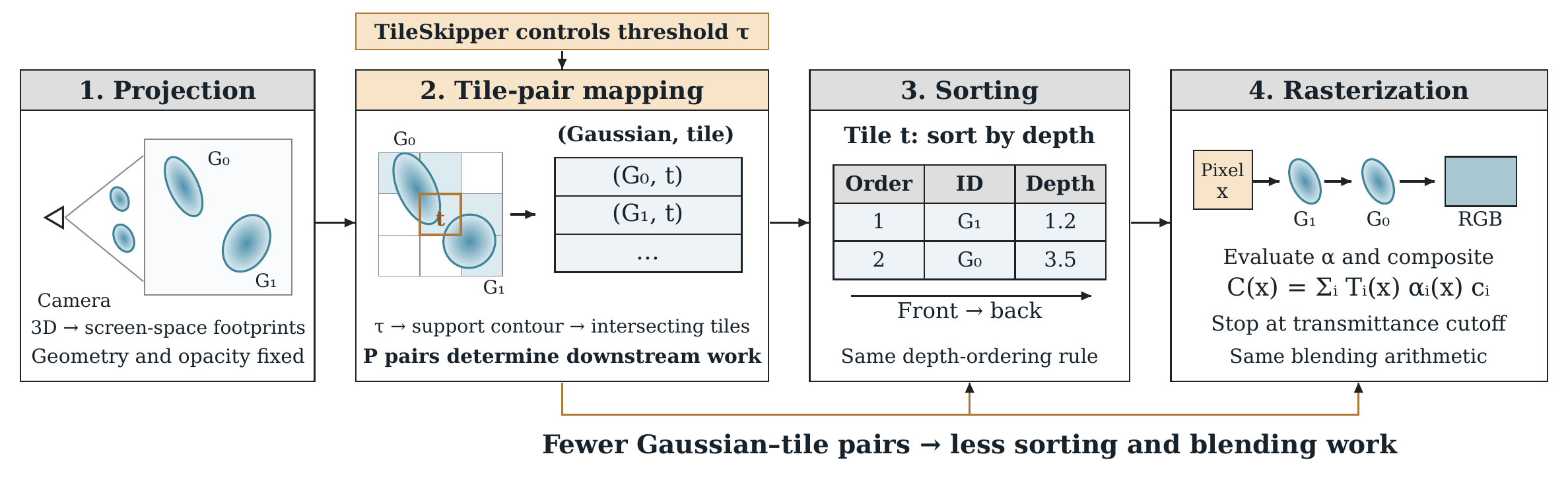}}
  \caption{\textbf{TileSkipper changes pair enumeration at stage~2.}
  The blue and orange outlines denote support under the baseline and compiled
  cutoffs. Tiles excluded by the compiled support generate no Gaussian--tile
  pair and therefore enter neither sorting nor blending. For tile $t$, the
  figure shows the resulting depth order and compositing expression; $T_i$ is
  transmittance before contribution $i$.}
  \label{fig:3dgs_pipeline}
\end{figure}

\paragraph{Rendering pipeline.} \Cref{fig:3dgs_pipeline} shows the four stages
relevant here. (1)~Each visible Gaussian is projected to a screen-space
footprint. (2)~Every intersecting tile is enumerated, creating one
(Gaussian, tile) pair per overlap. (3)~Each tile's assigned Gaussians are sorted
front-to-back. (4)~Pixels evaluate and alpha-blend the sorted Gaussians.
Vanilla 3DGS~\citep{kerbl20233dgs} uses a conservative $3\sigma$ box, whereas
Speedy-Splat~\citep{hanson2025speedysplat}, FastGS~\citep{ren2025fastgs}, and
SeeLe~\citep{huang2026seele} derive tile support from a fixed contribution
cutoff. TileSkipper changes only stage~2; Gaussian parameters, depth sorting,
and blending arithmetic remain unchanged.
At a fixed checkpoint and resolution, we empirically approximate frame latency
as $T\approx F+\beta\pairs$, where $\pairs$ is the Gaussian--tile pair count,
$F$ is the fitted pair-independent cost, and $\beta$ is the fitted latency
slope. This model explains why pair reduction need not translate proportionally
into runtime improvement; Appendix~\ref{sec:costmodel} gives the full model
and validation.

\subsection{Tile support from a scene-wide cutoff}
\label{sec:prelim}

Vanilla 3DGS~\citep{kerbl20233dgs} enumerates the $3\sigma$ box of the projected
Gaussian. We borrow the opacity level-set construction used by
AdR-Gaussian~\citep{wang2024adrgaussian}: threshold-based rasterizers bound the
$\thr$-level support of the splat's contribution. For
$0<\thr<\alpha_g$, the superlevel set
$\mathcal{S}_g(\thr)=\{x:\alpha_g(x)\ge\thr\}$ is an elliptical region. With
$\alpha_g(x)=\alpha_g\exp(-\tfrac12 r^2)$, its boundary
$\alpha_g(x)=\thr$ has squared Mahalanobis radius
\begin{equation}
r_g^2(\thr) \;=\; 2\ln\!\big(\alpha_g/\thr\big).
\label{eq:support_radius}
\end{equation}
This is the normalized form of AdR-Gaussian's adaptive-radius equation; the
level-set formula is prior work, not a TileSkipper
contribution. TileSkipper instead contributes the selection of heterogeneous,
lossy $\thr_g$ values from measured rendering risk. Raising $\thr$ shrinks the
footprint and removes pairs. Although every discarded individual contribution is
small, their cumulative image error has no such guarantee; this is why our
compiler validates complete candidate renders.

The implementations on which we build fix $\thr$ to one hand-chosen constant,
scene-independent and region-independent:

\begin{center}\small
\resizebox{\linewidth}{!}{%
\begin{tabular}{lll}
\toprule
Rasterizer & Where the constant lives & Value \\
\midrule
Speedy-Splat / AccuTile~\citep{hanson2025speedysplat} & tile-intersection test & $\base=1/255$ \\
FastGS~\citep{ren2025fastgs} & tile-intersection test & $\base=1/255$ \\
SeeLe~\citep{huang2026seele} & \texttt{lowest\_alpha\_coeff}$=\ln 255$ (compile-time) & $\base=1/255$ \\
\bottomrule
\end{tabular}}
\end{center}

A single constant must protect the most sensitive content represented by its
selection views. \Cref{app:local_tolerance} shows that locally safe thresholds
span $16\times$ within one image, establishing that a scene-global value can
leave slack in other regions. This diagnostic motivates heterogeneous control
but does not quantify its deployed benefit; the matched-quality comparison in
\Cref{tab:matched} measures that benefit directly. TileSkipper replaces the
constant with a compiled, static, per-Gaussian value while leaving checkpoint
parameters, sorting, and blending unchanged.

\subsection{Compiling a discrete tail policy}
\label{sec:compiler}

The policy assigns every Gaussian one multiplier from a fixed discrete ladder
$\ladder=\{1,1.05,1.1,1.15,1.25,1.5,2,3,4,6,8\}$, applied as
$\thr_g=m_g\base$ on the default Speedy-Splat path. If the threshold reaches
Gaussian opacity, that Gaussian produces no tile pairs; its checkpoint parameters
are retained. The FastGS port caps the threshold as detailed in
\Cref{app:threshold_semantics}. Choosing $m_g$ independently for millions of Gaussians is
separable for a fixed Lagrange multiplier but exposes millions of independent
decisions. Our default groups them into $\Kcl=64$ decisions as a regularized
search granularity, then exports the chosen level for every Gaussian. We compile
in five steps.

\paragraph{(1) Factorized probes.} For every Gaussian and ladder rung we collect
two quantities on calibration views. A global render at rung $\lvl$ records the
exact change in enumerated pairs and attributes it to each Gaussian,
$\save_{g,\lvl}$. A custom backward pass classifies pair removal with the same
tile predicate and accumulates
$\risk_{g,\lvl}=\sum_{v,x,c}(\alpha_{g,v}(x)\,\partial C_{v,c}(x)/\partial
\alpha_{g,v}(x))^2$ over pixels in pairs removed at rung $\lvl$. This is an
isolated-deletion RGB-energy statistic for removing one
Gaussian--tile contribution while holding the others fixed, but it omits cross
terms when many contributions are removed jointly. It is therefore a ranking
proxy, not an exact prediction of joint image error. Both tables require
$O(|\ladder|V)$ rasterizer passes rather than one render per Gaussian; the
held-out stage below validates their joint effect.

\paragraph{(2) Risk stratification.} At the most aggressive ladder rung,
Gaussians are bucketed into $S=8$ quantile strata by
$\log((\save_{g,\lvl}+\epsilon)/(\risk_{g,\lvl}+\epsilon))$. Gaussians whose
tails are cheap to remove can therefore share a decision even when far apart.

\paragraph{(3) Workload-balanced spatial partition.} Within each stratum we sort
by Morton code and cut into contiguous blocks of \emph{equal measured
tile-pair work}, giving $\Kcl$ regions in total. Balancing on work rather than on
volume prevents a few dense regions from consuming most of the decision budget:
a uniform axis-aligned grid has a region-work coefficient of variation of
$3.1$--$4.8$, which our partition reduces to $0.026$--$0.060$. Region membership
is frozen after this step. This balancing is a conditioning choice rather than a
claimed source of latency gain; \Cref{tab:partition_ablation} isolates its effect.

\paragraph{(4) Constrained search.} Aggregating the probes per region gives
$S_{\clust,\lvl}$ and $R_{\clust,\lvl}$. For each of $48$ penalties $\lambda$
spaced geometrically over the observed efficiency range, we take the independent
optimum
\begin{equation}
\lvl^\star_\clust(\lambda) \;=\; \argmax_{\lvl\in\ladder}\;\big(S_{\clust,\lvl}-\lambda\,R_{\clust,\lvl}\big),
\label{eq:lagrangian}
\end{equation}
which costs one pass over a $\Kcl\times|\ladder|$ table. The resulting proposals,
together with the uniform-$\lvl$ policies that constitute the \textsc{Ours-Global}
candidate set, form the candidate set. Including these uniform candidates lets
\textsc{Ours-Regional} fall back to a compiler-selected global cutoff when spatial
heterogeneity does not pay.

\paragraph{(5) Held-out selection.} Every candidate is rendered on a disjoint set
of training views. Feasibility requires the minimum per-view teacher PSNR against
the unmodified backbone to exceed a floor and the worst-view absolute p99
residual to be at most a fixed cap. The p99 statistic takes the maximum absolute
RGB-channel residual on a fixed $4\times$ spatial subsample.
Among feasible candidates we keep the one that removes the most pairs. Test
views are never used.

\paragraph{Export.} The selected policy is written as one \texttt{uint8} level
per Gaussian plus a lookup table of at most $256$ thresholds --- one byte per
Gaussian, under $1$\,KB of table. At render time the rasterizer reads the byte
and indexes the table inside the tile-enumeration loop. There is no per-frame
policy inference, no additional kernel, and no change to sorting or blending.
For the default $\Kcl=64$, compilation averages $9.0$\,s per scene and takes at
most $14.0$\,s, excluding checkpoint load (\Cref{ssec:exp_cost_detail}). It requires only
the checkpoint and the training views; the backward probe updates no parameter.

\section{Experiments} \label{sec:experiments}
We evaluate ten model/backend integrations on the same thirteen scenes,
then isolate policy granularity and compare with AdaGScale. Appendix experiments
cover region construction, calibration sensitivity, temporal stability, transfer,
and systems costs.

\subsection{Setup} \label{ssec:exp_setup}
\paragraph{Datasets.} We use the standard 3DGS protocol on thirteen scenes: the nine MipNeRF360 scenes~\citep{barron2022mipnerf360} (\emph{bicycle}, \emph{flowers}, \emph{garden}, \emph{stump}, \emph{treehill}, \emph{bonsai}, \emph{counter}, \emph{kitchen}, \emph{room}), two Tanks~\&~Temples scenes~\citep{knapitsch2017tnt} (\emph{train}, \emph{truck}), and two Deep Blending scenes~\citep{hedman2018db} (\emph{drjohnson}, \emph{playroom}). We use the official train/test splits and the standard image downsampling used by 3DGS. Standard-resolution test images range from $979\!\times\!545$ to $1559\!\times\!1039$ ($0.53$--$1.62$ megapixels); the 4K setting rescales the same views to $3840$ pixels wide while preserving aspect ratio.

\paragraph{Backbones and thresholds.} Vanilla 3DGS and Mini-Splatting use the
same vanilla-compatible CUDA rasterizer; Mini-Splatting changes the surviving
Gaussian set but retains the standard PLY representation. Their released
$3\sigma$ tile box is replaced by the prior exact-$1/255$ bound before our
compiled policy is applied. Speedy-Splat is the canonical
fixed-threshold AccuTile backbone ($\base=1/255$). FastGS combines the same
cutoff with its native compact-box rasterizer. We transfer a Speedy-Splat policy
without recalibration into the native SeeLe and FlashGS rasterizers. 2DGS uses
surfel--ray intersection, while Scaffold-GS generates view-dependent neural
Gaussians from anchors; both released rasterizers enumerate a $3\sigma$ tile
box, so their drop-in ports first supply the exact $1/255$ bound. We separate
that prior-work transition from our compiler contribution in
\Cref{tab:backbone_decomposition,tab:fragment_backbones}. Finally, LightGaussian and Taming-3DGS test
pure checkpoints through the same AccuTile backend; the latter retains its
native absolute-opacity activation.

\paragraph{Metrics and timing protocol.} We report PSNR~\citep{wang2009mse}, SSIM~\citep{wang2004ssim}, LPIPS~\citep{zhang2018lpips}, the
(Gaussian, tile) pair ratio $R$ against the unmodified backbone, and average
per-frame render time on one NVIDIA RTX~5000 Ada. We distinguish ground-truth
PSNR from \emph{teacher PSNR}. The former compares a rendered image with the
dataset image; SSIM and LPIPS use the standard 3DGS evaluation code, with
VGG features for LPIPS, and are averaged over the same test views. For an approximate rendering $I$ and the unmodified renderer
$I_0$, both on the $[0,1]$ RGB scale, teacher PSNR is
\[
 P_{\rm teacher}=-10\log_{10}\!\left(
 \operatorname*{mean}_{x,c}(I(x,c)-I_0(x,c))^2\right).
\]
Thus the 60, 55, and 50\,dB operating points correspond to teacher-MSE targets
of $10^{-6}$, $10^{-5.5}$, and $10^{-5}$, respectively; 50\,dB permits
$3.16\times$ the approximation MSE of 55\,dB. These targets do not denote PSNR
against the dataset images. Every comparison is paired
within one backbone and one process; arms are rotated, synchronized per frame,
and use identical caching. Speedy-Splat caches its frozen per-Gaussian
activations once in every arm. The 2DGS timing includes ray--surfel intersection;
the Scaffold-GS timing includes its per-frame MLP Gaussian generation, which our
policy cannot reduce. Image writing is disabled. Speedups are geometric means
over scenes; cross-row absolute latency is not compared.

\paragraph{Compiler settings.} Unless stated otherwise we use the ladder
$\ladder=\{1,1.05,1.1,1.15,1.25,1.5,2,3,4,6,8\}$, $S=8$ risk strata,
$\Kcl=64$ workload-balanced regions, $48$ Lagrangian penalties,
and $16$ calibration plus $32$ disjoint selection views drawn from the training
split. Feasibility requires a minimum-view teacher PSNR of $48$\,dB and a
worst-view absolute p99 residual of at most $0.012$ on the $[0,1]$ RGB scale.
The teacher is the unmodified base path for each row, except that vanilla 3DGS,
Mini-Splatting, 2DGS, and Scaffold-GS use their exact-$1/255$ transition. Test views are never used for
calibration or selection. The probe uses backward-mode differentiation but no
checkpoint parameter or optimizer state is updated.

\subsection{Main results} \label{ssec:exp_main}

\Cref{tab:main} reports deployment-facing ``base versus base + Ours''
comparisons, while the accompanying decomposition tables identify which portion
is attributable to TileSkipper. On the six paths that already expose an
opacity-aware tile bound, the reported gains are compiler-only:
$1.096\times$ on Speedy-Splat, $1.009\times$ on FastGS,
$1.091\times$ on SeeLe, $1.121\times$ on FlashGS,
$1.095\times$ on LightGaussian, and $1.051\times$ on Taming-3DGS.
The vanilla 3DGS, Mini-Splatting, 2DGS, and Scaffold-GS rows instead include a
prior $3\sigma\!\rightarrow\!1/255$ transition. For example, the
$1.493\times$ vanilla-3DGS total decomposes into a $1.454\times$ exact-bound
transition and a further $1.027\times$ compiler gain; the corresponding
Mini-Splatting factors are $1.266\times$ and $1.063\times$.
\Cref{tab:backbone_decomposition,tab:fragment_backbones} gives the full
accounting. Across all ten integrations, the mean SSIM change is between
$-3.8\!\times\!10^{-5}$ and $-2.5\!\times\!10^{-4}$, while LPIPS changes
by $-3.1\!\times\!10^{-5}$ to $-3.9\!\times\!10^{-4}$ (lower is better).

The 2DGS and Scaffold-GS ports improve by $1.474\times$ and $1.320\times$.
The larger 2DGS gain reflects expensive ray--surfel work per removed pair. On
Scaffold-GS, the rasterizer itself improves by $2.10\times$, but the per-frame MLP
occupies $35$--$65\%$ of latency and limits the end-to-end result. Because both
released paths start from a $3\sigma$ box, we expose rather than claim their
exact-bound contribution in \Cref{tab:backbone_decomposition}.

The gain grows with resolution in the measured regime
(\Cref{tab:resolution_sweep} and \Cref{fig:dispatch}a). Stage profiles show the
duplicate-plus-sort share increasing from $21.4\%$ to $34.2\%$ on
\emph{bicycle} and from $31.6\%$ to $41.6\%$ on \emph{counter}; thresholding
also changes correlated blending work. Additional
ablations, transfer breakdowns, temporal tests, and cost-model diagnostics are
reported in the appendix.

\begin{table}[t]
\caption{\textbf{Base model/backend versus the same path plus Ours ($\Kcl=64$).} Every row has thirteen scenes and twelve balanced same-process timing cycles. All quality deltas are Ours minus the named base (higher PSNR/SSIM and lower LPIPS are better). $R$ is the mean pair ratio; intervals bootstrap scenes and parentheses count wins. SeeLe and FlashGS transfer a Speedy-Splat policy unchanged into the named native rasterizer. LightGaussian and Taming-3DGS vary the checkpoint on one common AccuTile backend; Taming's native absolute-opacity activation is preserved. The vanilla 3DGS, Mini-Splatting, 2DGS, and Scaffold-GS rows include the prior exact-bound transition, separated in \Cref{tab:backbone_decomposition,tab:fragment_backbones}.}
\label{tab:main}
\centering\small
\setlength{\tabcolsep}{4pt}
\resizebox{\linewidth}{!}{%
\begin{tabular}{llrrrrrl}
\toprule
Base model/backend & Integration & Scenes & $\Delta$PSNR & $\Delta$SSIM & $\Delta$LPIPS$\downarrow$ & $R$ & +Ours speedup [95\% CI] (wins) \\
\midrule
\multicolumn{8}{l}{\emph{Native and native-compatible rasterizer integrations}} \\
vanilla 3DGS~\citep{kerbl20233dgs} & 3DGS CUDA & 13 & $-0.003$ & $-0.000069$ & $-0.000092$ & $0.330$ & $\mathbf{1.493\times}$ $[1.350,1.656]$ (13/13) \\
Mini-Splatting~\citep{fang2024minisplatting} & 3DGS CUDA & 13 & $-0.008$ & $-0.000104$ & $-0.000172$ & $0.435$ & $\mathbf{1.346\times}$ $[1.260,1.451]$ (13/13) \\
Speedy-Splat~\citep{hanson2025speedysplat} & native & 13 & $-0.007$ & $-0.000109$ & $-0.000202$ & $0.825$ & $\mathbf{1.096\times}$ $[1.078,1.116]$ (13/13) \\
FastGS~\citep{ren2025fastgs} & native & 13 & $-0.004$ & $-0.000088$ & $-0.000098$ & $0.979$ & $\mathbf{1.009\times}$ $[1.005,1.014]$ (11/13) \\
SeeLe~\citep{huang2026seele} & native transfer & 13 & $-0.008$ & $-0.000120$ & $-0.000073$ & $0.800$ & $\mathbf{1.091\times}$ $[1.070,1.115]$ (13/13) \\
FlashGS~\citep{feng2025flashgs} & native transfer & 13 & $-0.005$ & $-0.000254$ & $-0.000386$ & $0.821$ & $\mathbf{1.121\times}$ $[1.098,1.145]$ (13/13) \\
2DGS~\citep{huang20242dgs} & native & 13 & $-0.008$ & $-0.000174$ & $-0.000091$ & $0.641$ & $\mathbf{1.474\times}$ $[1.370,1.586]$ (13/13) \\
Scaffold-GS~\citep{lu2024scaffold} & native & 13 & $-0.008$ & $-0.000132$ & $-0.000115$ & $0.408$ & $\mathbf{1.320\times}$ $[1.218,1.443]$ (13/13) \\
\addlinespace
\multicolumn{8}{l}{\emph{Checkpoint integrations on a common AccuTile backend}} \\
LightGaussian~\citep{fan2024lightgaussian} & AccuTile & 13 & $-0.006$ & $-0.000123$ & $-0.000196$ & $0.834$ & $\mathbf{1.095\times}$ $[1.072,1.118]$ (13/13) \\
Taming-3DGS~\citep{mallick2024taming3dgs} & AccuTile & 13 & $-0.001$ & $-0.000038$ & $-0.000031$ & $0.926$ & $\mathbf{1.051\times}$ $[1.032,1.073]$ (13/13) \\
\bottomrule
\end{tabular}}
\end{table}

\subsection{What regional allocation adds}
\label{ssec:exp_matched}

\Cref{tab:matched} compares Global ($\Kcl=1$), Regional ($\Kcl=64$),
and per-Gaussian control using the same probes. Each family's measured curve
is interpolated linearly in teacher MSE to 60/55/50\,dB; these analytical points
need not correspond to a single exported policy. Regional is $4.0$--$4.6\%$
faster than Global at native resolution and up to $8.3\%$ faster at 4K, but
ties per-Gaussian control ($0.986$--$1.007\times$). Thus heterogeneous allocation
improves over scene-level calibration, while region sharing has no demonstrated
speed or storage advantage over per-Gaussian decisions. Both export one byte
per Gaussian; Appendix~\ref{app:comparison_details} gives the full protocol.

\begin{table}[t]
\caption{Strict matched-quality ablation of compiler granularity on thirteen scenes. Each variant's own (distortion, latency) Pareto envelope is interpolated at a common teacher-PSNR target against the unmodified backbone. \textsc{Ours-Global} is our compiler with one decision ($\Kcl=1$), not an external baseline; Regional is the full $\Kcl=64$ method. Brackets are $95\%$ scene-bootstrap intervals. Regional improves on \textsc{Ours-Global} everywhere and \emph{ties} the per-Gaussian variant. Thus $\Kcl$ is a decision regularizer and search granularity, not a demonstrated source of speed or storage reduction.}
\label{tab:matched}
\centering\small
\setlength{\tabcolsep}{4pt}
\begin{tabular}{llrrrr}
\toprule
Resolution & Teacher PSNR & \shortstack{Regional /\\Ours-Global} & \shortstack{Per-G /\\Ours-Global} & \shortstack{Regional /\\per-G} & Wins \\
\midrule
Standard & 60\,dB & $\mathbf{1.0397\times}$ {\scriptsize$[1.0290,1.0508]$} & $1.0435\times$ & $0.9963\times$ & 13/13 \\
         & 55\,dB & $\mathbf{1.0457\times}$ {\scriptsize$[1.0373,1.0551]$} & $1.0460\times$ & $0.9997\times$ & 13/13 \\
         & 50\,dB & $\mathbf{1.0401\times}$ {\scriptsize$[1.0313,1.0504]$} & $1.0332\times$ & $1.0067\times$ & 13/13 \\
\midrule
4K       & 60\,dB & $\mathbf{1.0413\times}$ {\scriptsize$[1.0208,1.0612]$} & $1.0556\times$ & $0.9864\times$ & 10/13 \\
         & 55\,dB & $\mathbf{1.0709\times}$ {\scriptsize$[1.0500,1.0923]$} & $1.0811\times$ & $0.9906\times$ & 13/13 \\
         & 50\,dB & $\mathbf{1.0825\times}$ {\scriptsize$[1.0632,1.1029]$} & $1.0816\times$ & $1.0008\times$ & 13/13 \\
\bottomrule
\end{tabular}
\end{table}

\subsection{Closest-method comparison}
\Cref{tab:adagscale_official} compares the two standalone methods under the
same training-view quality constraints at native and 4K resolution. These
constraints do not imply identical attained errors. At native resolution,
the pooled $1.010\times$ speedup has an interval of $[0.990,1.025]$;
this does not establish broad runtime dominance. At 4K, the pooled speedup is
$1.076\times$ $[1.047,1.102]$, with 12/13 scene wins.
Dataset-specific results and all three quality metrics are reported in the
same table; \Cref{ssec:ada_common_resolution} details the protocol.

\begin{table}[t]
\centering
\caption{\textbf{Standalone comparison with AdaGScale at native and 4K resolution.}
Both methods use the same Speedy-Splat checkpoints and AdaGScale FlashGS renderer,
with policies selected under common training-view quality constraints
(details in \Cref{ssec:ada_common_resolution}).
Deltas are TileSkipper minus AdaGScale; $R$ is the TileSkipper/AdaGScale pair ratio,
and speedup is AdaGScale/TileSkipper latency.
Brackets are 95\% scene-bootstrap intervals; parentheses count speed wins.
The two-scene intervals are descriptive.}
\label{tab:adagscale_official}
\small
\resizebox{\linewidth}{!}{%
\begin{tabular}{lrrrrrl}
\toprule
Dataset & Scenes & $\Delta$PSNR & $\Delta$SSIM & $\Delta$LPIPS$\downarrow$ & $R$ & TileSkipper speedup [95\% CI] (wins) \\
\midrule
\multicolumn{7}{c}{\textbf{(a) Native resolution}} \\
\midrule
All scenes & 13 & $+0.0059$ & $+0.000021$ & $-0.000337$ & $0.945$ & $1.010\times$ $[0.990,1.025]$ (10/13) \\
\midrule
Mip-NeRF 360 & 9 & $+0.0056$ & $+0.000066$ & $-0.000349$ & $0.934$ & $1.017\times$ $[1.004,1.030]$ (7/9) \\
Tanks \& Temples & 2 & $-0.0001$ & $-0.000032$ & $-0.000161$ & $1.008$ & $0.960\times$ $[0.920,1.002]$ (1/2) \\
Deep Blending & 2 & $+0.0138$ & $-0.000126$ & $-0.000461$ & $0.930$ & $1.027\times$ $[1.024,1.029]$ (2/2) \\
\midrule
\multicolumn{7}{c}{\textbf{(b) 4K}} \\
\midrule
All scenes & 13 & $+0.0366$ & $+0.000151$ & $-0.000342$ & $0.929$ & $1.076\times$ $[1.047,1.102]$ (12/13) \\
\midrule
Mip-NeRF 360 & 9 & $+0.0460$ & $+0.000195$ & $-0.000375$ & $0.917$ & $1.085\times$ $[1.061,1.113]$ (9/9) \\
Tanks \& Temples & 2 & $+0.0084$ & $+0.000046$ & $-0.000044$ & $0.986$ & $1.016\times$ $[0.952,1.084]$ (1/2) \\
Deep Blending & 2 & $+0.0227$ & $+0.000060$ & $-0.000492$ & $0.924$ & $1.096\times$ $[1.086,1.106]$ (2/2) \\
\bottomrule
\end{tabular}}
\end{table}

\section{Conclusion}

Across ten model/backend integrations, a scene-wide tile-footprint threshold
can be compiled into a heterogeneous static policy rather than hand-tuned. The
one-byte-per-Gaussian result performs no training or per-frame policy inference
and composes with compact checkpoints and independently optimized rasterizers.
Detailed limitations and failure cases appear in \Cref{ssec:exp_limitations}.

\section*{Reproducibility Statement}
The method section describes the policy construction and selection procedure.
The experiments section specifies the datasets, evaluation metrics, compiler
settings, and timing protocol. The appendices provide additional details on
backend integration, threshold semantics, and experimental protocols to support
reproduction of the reported results. We will publicly release the source code
and evaluation scripts upon acceptance.

\section*{AI Use Statement}
Generative AI tools assisted with language polishing. The authors take full responsibility for the entire content of this paper.

\bibliographystyle{plainnat}
\bibliography{references}

\newpage
\appendix
\section{Related-work comparison}
\label{app:related_table}
\begin{table}[h]
                        \centering\small
                        \caption{Positioning by rasterization stage and decision unit. TileSkipper compiles heterogeneous, static per-Gaussian cutoffs without checkpoint updates or a single hand-set scene-wide cutoff; complete held-out renders validate each proposed policy.}
                        \label{tab:related}
                        \resizebox{\linewidth}{!}{%
\begin{tabular}{llll}
                        \toprule
                        Family & Stage & Control unit & Parameter updates \\
                        \midrule
                        Speedy-Splat (AccuTile) / FastGS~\citep{hanson2025speedysplat,ren2025fastgs} & Tile-bounding & Whole scene, hand-set & --- \\
                        AdR-Gaussian / FlashGS / QuadBox~\citep{wang2024adrgaussian,feng2025flashgs,li2026quadbox} & Tile geometry & Per projected Gaussian & --- \\
                        AdaGScale~\citep{jo2026adagscale} & Tile-bounding & Per view, per Gaussian & --- \\
                        SeeLe~\citep{huang2026seele} & View prefetch + blending & Camera cluster & fine-tuning \\
                        Fragment Pruning~\citep{ye2024fragmentpruning} & Per-pixel rasterization & Per-Gaussian & learned \\
                        Gaussian-set pruning~\citep{lee2024compact3d,girish2024eagles,niemeyer2025radsplat,hanson2025pup3dgs} & Gaussian set & Per-Gaussian & varies \\
                        View-adaptive scaling~\citep{yu2024mipsplatting} & Footprint shape & Per-view & --- \\
                        \textbf{TileSkipper (ours)} & Tile-bounding & \textbf{Compiled per-Gaussian} & \textbf{none} \\
                        \bottomrule
                        \end{tabular}
}
                        \end{table}

\section{Cost model and optional pre-culling}
\label{app:systems_details}
These diagnostics explain the measured scope of pair reduction and the optional
pre-culling extension. The core threshold policy in \Cref{sec:compiler} can be
used without pre-culling.

\subsection{A three-term cost model}
\label{sec:costmodel}

\begin{figure}[t]
  \centering
  \includegraphics[width=\linewidth]{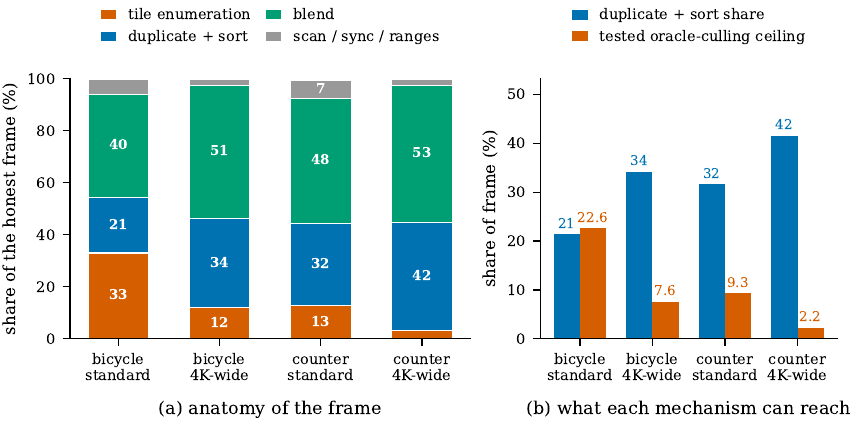}
  \caption{\textbf{Measured headroom under two controlled interventions.}
  \emph{(a)} Stage shares measured with the rasterizer's own CUDA events. \emph{(b)} The mechanisms
  have different primary targets, but their runtime effects are not assumed to be disjoint or
  additive. Thresholding changes tile enumeration, duplicate/sort work, and correlated blending;
  pre-culling avoids processing rejected Gaussian blocks and may also change downstream work.
  Within each profiled scene, the duplicate-plus-sort share grows with resolution:
  $21.4\%\!\rightarrow\!34.2\%$ for \emph{bicycle} and
  $31.6\%\!\rightarrow\!41.6\%$ for \emph{counter}. The estimated ceiling of the tested
  perfect, free visibility culler falls from $22.6\%$ to $7.6\%$ and from $9.3\%$ to
  $2.2\%$, respectively. Dispatch uses measured net benefit, including planning overhead.}
  \label{fig:anatomy}
\end{figure}

Following the Gaussian-, pair-, and pixel-level cost decomposition of
AdR-Gaussian~\citep{wang2024adrgaussian}, we instantiate a measured affine model
for the rasterizers studied here; the decomposition itself is borrowed rather
than claimed as a contribution. Let $T$ denote mean end-to-end frame latency,
$\Npts$ the number of Gaussians processed per frame, $\pairs$ the number of
enumerated Gaussian--tile pairs, and $|\Omega|$ the number of pixels. At a fixed
checkpoint and resolution, a global-cutoff sweep changes $\pairs$ while
$\Npts$ and $|\Omega|$ remain fixed, so we first fit the effective relation
$T=F+\beta\pairs$. Here $\beta\pairs$ denotes all latency that co-varies with
pair count in this controlled sweep, including enumeration, sorting, and
correlated downstream rasterization work; it is not a claim that every pair
has an identical or context-independent cost. Separate visible-subset and
resolution measurements model the remaining variation with $\Npts$ and
$|\Omega|$, yielding the empirical affine model
\begin{equation}
T \;=\; \underbrace{a\,\Npts}_{\text{per-Gaussian}} \;+\; \underbrace{\beta\,\pairs}_{\text{per-pair}} \;+\; \underbrace{c\,|\Omega|}_{\text{per-pixel}} .
\label{eq:costmodel}
\end{equation}
Coefficients are specific to the measured rasterizer, resolution, policy
family, and GPU. We use the model for interpolation and dispatch within those
controlled sweeps, rather than as a causal decomposition or a
hardware-independent law. \Cref{fig:anatomy} visualizes the measured stage
shares and the terms addressed by thresholding and pre-culling.

\paragraph{Pair count predicts latency within cutoff sweeps.} Across eight
scene-global cutoff settings in each of nine diagnostic
scene--backbone--resolution cells, the linear fit gives
$R^2=0.9934$--$0.9988$, where $R^2$ is the coefficient of determination.
Thus pair count explains $99.34\%$--$99.88\%$ of the observed latency variance
within these fixed-policy-family sweeps. This supports exactly counted pair
removal as a one-dimensional workload surrogate for selecting cutoff policies
in the measured regime. It does not establish equal cost per pair or
sufficiency of pair count across different policy families, checkpoint
densities, or pre-culling configurations.

\paragraph{The per-Gaussian term is invisible to any threshold policy.}
A cutoff sweep leaves $\Npts$ fixed and therefore cannot reduce per-Gaussian
projection and preprocessing. On \emph{bicycle} at native resolution, the
preprocessing time has a fitted slope of $-0.023$\,ns per pair and $R^2=0.26$.
The low $R^2$ means that pair count explains only $26\%$ of this stage's small
timing variation; it is not a hypothesis test that the slope equals zero. The
fitted pair-independent preprocessing intercept is $35.1\%$ of baseline frame
latency on that checkpoint.

\paragraph{Resolution and density move the measured balance.} Increasing
resolution changes both $\pairs$ and $|\Omega|$ while leaving $\Npts$ fixed. In
the two scenes with stage profiles, the duplicate-plus-sort share grows from
$21.4\%$ to $34.2\%$ for \emph{bicycle} and from $31.6\%$ to $41.6\%$ for
\emph{counter}; we do not mix the two scenes into one resolution estimate.
Increasing checkpoint density can change both $\Npts$ and $\pairs$, so Gaussian
count alone is not a dispatch rule. We instead dispatch from the measured share
$a\Npts/T$.

\subsection{Conservative region pre-culling}
\label{sec:precull}

A threshold policy cannot prevent a Gaussian from entering projection and
preprocessing. We therefore add a complementary mechanism that skips
conservatively rejected Gaussian blocks before projection; its measured image
effect is bit-exact rather than error-budgeted.

At export time the model is reordered once by Morton code and cut into
fixed-size contiguous blocks, so that spatially adjacent Gaussians are adjacent
in memory. For each block we precompute one axis-aligned box that contains the
full $\base$-support of every Gaussian it holds, inflated by a safety factor.
Each frame, the $\Kcl'$ block boxes are tested against the camera frustum and the
surviving blocks are handed to the rasterizer as contiguous ranges; Gaussians in
rejected blocks are never read, so they cost nothing in projection, prefix scan
or duplication.

The inflated support boxes are designed to be conservative with respect to the
backbone's finite-support predicate. Empirically, the resulting images are
bit-identical to the unmodified backbone in all 52 evaluated runs. We therefore
call the measured implementation image-bit-exact, while avoiding a claim of
numerical exactness for arbitrary cameras or hardware. Planning time is included
in every reported pre-culling timing.

\subsection{Dispatch: which mechanism, and when}
\label{sec:dispatch}

The mechanisms have different primary targets, but their measured savings need
not be disjoint or additive. We therefore use \Cref{eq:costmodel} as an
empirical dispatch model rather than as a causal partition of runtime. Fitting
$a$ from one oracle visible-subset comparison provides an estimated ceiling for
the tested pre-culling design at that checkpoint, resolution, and hardware.
Across thirteen scenes, this estimate matches the realized pre-culling ceiling
to a mean absolute error of $0.004$.

We enable pre-culling only when its predicted removable time exceeds its measured
planning overhead. In our standard-resolution sweep, $a\Npts/T>20\%$ is a reliable
``on'' regime and values below $10\%$ are reliably ``off''; intermediate cases
use the measured inequality rather than a Gaussian-count heuristic. The policy
itself remains enabled in both cases. \Cref{sec:experiments} reports both arms.

\section{Additional comparison protocols and resolution results}
\label{app:comparison_details}

\subsection{Compiler granularity protocol}
\Cref{tab:fixed} compares against constants a practitioner might plausibly pick.
\Cref{tab:matched} instead compares three granularities of our compiler.
\textsc{Ours-Global} ($\Kcl=1$) automatically selects one scene-level threshold
under the same held-out constraints; \textsc{Ours-Regional} ($\Kcl=64$) and the
per-Gaussian variant reuse the same measurements but expose more decisions. For each family we build
its own $(\text{distortion},\text{latency})$ Pareto envelope and read off latency at a
common teacher-PSNR target, so the arms are compared at equal distortion rather than at
equal work. The 60/55/50\,dB labels are these teacher-distortion targets. For
each method, we interpolate linearly in teacher MSE between adjacent measured
policies; the interpolated point summarizes its measured curve and need not
coincide with one exported discrete policy.

The full regional compiler improves on \textsc{Ours-Global} at every target and on
$13/13$ scenes at standard dataset resolution, by $4.0$--$4.6\%$, rising to $8.3\%$ at 4K.
This is the incremental contribution of heterogeneous allocation, not the total
compiler gain reported against Official-Fixed in \Cref{tab:main}.

The same table carries an important negative result. Against a per-Gaussian policy
compiled from the same probes, region sharing is a tie: $0.986$--$1.007\times$ across
three targets and two resolutions. The per-Gaussian Lagrangian search is separable and
costs $O(\Npts|\ladder|)$ per penalty, not $|\ladder|^{\Npts}$. Regional sharing reduces
the number of independently selected decisions to $\Kcl|\ladder|$ and acts as a
regularizer, but both implementations currently export one level byte per Gaussian.
Thus our evidence supports compilation and heterogeneous control, not a storage or speed
advantage of $\Kcl=64$ over per-Gaussian control. $\Kcl=1$, $64$, and $\Npts$ in
\Cref{tab:matched} expose that policy-granularity spectrum directly.

\subsection{AdaGScale comparison protocol}
\label{ssec:ada_common_resolution}
For \Cref{tab:adagscale_official}, AdaGScale recalibrates its twenty-bin
$T_{\rm upper}$ table for each checkpoint and resolution on sixteen training
views (per-bin maxima; empty bins set to one), then searches $K_{\mathrm{Ada}}
\in[0,64]$ in eighteen steps. TileSkipper re-evaluates its frozen native-resolution
$\Kcl=64$ candidate family. Both select on the same 32 held-out training views,
requiring minimum-view teacher PSNR $\ge48$\,dB and worst-view p99 $\le0.012$
relative to the same renderer at cutoff $1/255$.
Here p99 is the 99th percentile of per-pixel maximum absolute RGB error on a
stride-four grid, maximized over views. These constraints do not enforce equal
attained quality. Policies are frozen before evaluation on all test views;
timing uses twelve balanced same-process cycles with per-frame CUDA synchronization.

The 4K setting uses 3840-pixel width with the original aspect ratio and
consistently scaled camera intrinsics. Reference images are Lanczos-resized
from the native images, so this setting measures increased rendering workload,
not newly captured image detail.

\section{Additional experiments}

\begin{figure}[t]
  \centering
  \includegraphics[width=\linewidth]{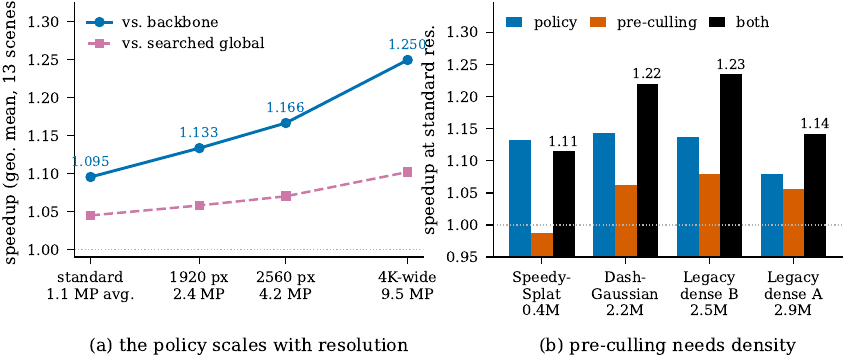}
  \caption{\textbf{Each mechanism has a regime, and the cost model names it.} \emph{(a)} The
  compiled policy strengthens with resolution; both $\pairs$ and $|\Omega|$ change in this
  sweep. The separate stage profiles in \Cref{fig:anatomy} show the within-scene changes and
  do not conflate the \emph{bicycle} and \emph{counter} endpoints. \emph{(b)}
  Pre-culling strengthens as the measured $a\Npts/T$ share grows. The dense bars are stress-test
  checkpoints; the two legacy packaged checkpoints are not used as evidence for any named model
  family. On Speedy-Splat's $0.4$M checkpoint the frustum test costs more than it saves and
  ``both'' is \emph{worse} than the policy alone, which is the case the dispatch rule catches.}
  \label{fig:dispatch}
\end{figure}

\subsection{Fixed-policy resolution stress test}
\label{ssec:resolution_sweep}
\Cref{tab:resolution_sweep} reports the source of the resolution numbers quoted
in the abstract and introduction. It uses one timing protocol consistently at
all widths and applies the same held-out-selected policy without recalibration.
The standard-resolution geometric mean, $1.0955\times$, agrees with the
independent twelve-cycle main-table retiming, $1.096\times$. We report both the equally weighted dataset macro and the thirteen-scene
geometric mean so that the abstract's $1.088/1.238\times$ paired comparison is
directly traceable. Because the higher-resolution inputs
are resampled from the same test views, this experiment establishes scaling
with raster workload rather than performance on newly captured 4K detail.

\begin{table}[t]
\caption{\textbf{Fixed-policy resolution stress test on AccuTile, all thirteen
scenes.} The same deployable $\Kcl=64$ policies, selected using held-out training
views at standard resolution, are applied unchanged at every width; test views
are used only for evaluation. Higher-resolution rows bilinearly resample the
same source views and therefore test workload scaling rather than natively
captured high-resolution content. Frozen per-Gaussian attributes are cached once
in both arms. Timing rotates AccuTile, \textsc{Ours-Global}, and
\textsc{Ours-Regional} in one process with three warmup and seven measured
repetitions per view. Speedup is AccuTile/Regional latency; ``dataset macro''
equally weights the three dataset-level arithmetic means, while ``scene geom.''
is the geometric mean over all thirteen scenes. $R$ and all quality deltas are
arithmetic scene means over full official test sets; deltas are Regional minus
AccuTile (higher PSNR/SSIM and lower LPIPS are better).}
\label{tab:resolution_sweep}
\centering\small
\setlength{\tabcolsep}{4pt}
\resizebox{\linewidth}{!}{%
\begin{tabular}{lrrrrrr}
\toprule
Width & MP/frame & $R$ & $\Delta$PSNR & $\Delta$SSIM & $\Delta$LPIPS$\downarrow$ & Speedup (dataset macro / scene geom.) \\
\midrule
Standard & $1.15$ & $0.8248$ & $-0.0074$ & $-0.000109$ & $-0.000202$ & $1.0879\times / 1.0955\times$ \\
$1920$ px & $2.38$ & $0.8044$ & $-0.0141$ & $-0.000416$ & $-0.000145$ & $1.1283\times / 1.1334\times$ \\
$2560$ px & $4.23$ & $0.7908$ & $-0.0166$ & $-0.000532$ & $-0.000167$ & $1.1615\times / 1.1665\times$ \\
$3840$ px & $9.51$ & $0.7750$ & $-0.0231$ & $-0.000742$ & $-0.000221$ & $1.2384\times / 1.2495\times$ \\
\bottomrule
\end{tabular}}
\end{table}

\subsection{Region-count ablation on Speedy-Splat}
\label{ssec:exp_k}

\Cref{tab:k_ablation} sweeps $\Kcl$ from $1$ to $1024$ on the same
thirteen Speedy-Splat scenes, using the existing construction rule with
$S_{\mathrm{eff}}=\min(8,\Kcl)$ risk strata.
The $\Kcl=1$ compiler already gives $1.048\times$ over the official fixed cutoff.
The added $\Kcl=2$ and $4$ policies yield $1.0165\times$ and $1.0351\times$
over their same-process $\Kcl=1$ controls; $\Kcl=8$ yields $1.0412\times$.
A small number of groups therefore captures most of the observed benefit.
For $\Kcl=64$, $128$, and $256$, the corresponding ratios are
$1.0472\times$, $1.0468\times$, and $1.0451\times$.
The added $\Kcl=512$ and $1024$ rows reach $1.0488\times$ and
$1.0519\times$, with mean pair ratios of $0.8158$ and $0.8164$ and mean
compilation times of $11.2$ and $15.0$ seconds, compared with $9.0$ seconds
at $\Kcl=64$. Each ratio uses its own same-process global control;
differences between separately timed regional configurations are not direct
paired comparisons. These are common-constraint results rather than exact
quality matching. We retain the globally fixed $\Kcl=64$ rather than selecting
the best $\Kcl$ per scene. This independent sweep's
$\Kcl=64$ speedup ($1.095\times$) agrees with the independently repeated main
operating-point retiming ($1.096\times$).

\begin{table}[t]
\caption{\textbf{Region-count ablation on Speedy-Splat, all thirteen scenes.} $\Kcl=1$ is our compiler restricted to one scene-level decision; larger $\Kcl$ increases the number of policy groups, with spatial workload balancing within each risk stratum. Every row uses the same ladder, calibration/selection views, and quality constraints; the existing construction uses $S_{\mathrm{eff}}=\min(8,\Kcl)$ risk strata. Latency uses the same honest-cache, rotating three-arm protocol as \Cref{tab:main}; brackets are $95\%$ scene-bootstrap intervals. We retain the globally fixed $\Kcl=64$; the added $\Kcl=2,4,512,1024$ rows use the same three-arm evaluation protocol.}
\label{tab:k_ablation}
\centering\small
\setlength{\tabcolsep}{4pt}
\resizebox{\linewidth}{!}{%
\begin{tabular}{rrrrrr}
\toprule
$\Kcl$ & mean $\Delta$PSNR & $R$ & vs. Official-Fixed & vs. $\Kcl=1$ [95\% CI] & compile (s) \\
\midrule
$1$ & $-0.0041$ & $0.9192$ & $1.0484\times$ & $1.0000\times$ & $2.8$ \\
$2$ & $-0.0047$ & $0.8868$ & $1.0641\times$ & $1.0165\times$ $[1.0105,1.0223]$ & $3.8$ \\
$4$ & $-0.0051$ & $0.8479$ & $1.0843\times$ & $1.0351\times$ $[1.0276,1.0424]$ & $4.9$ \\
$8$ & $-0.0067$ & $0.8329$ & $1.0894\times$ & $1.0412\times$ $[1.0298,1.0525]$ & $7.0$ \\
$16$ & $-0.0054$ & $0.8306$ & $1.0921\times$ & $1.0422\times$ $[1.0289,1.0551]$ & $7.7$ \\
$32$ & $-0.0064$ & $0.8273$ & $1.0949\times$ & $1.0436\times$ $[1.0316,1.0555]$ & $8.7$ \\
$\mathbf{64}$ & $-0.0080$ & $0.8227$ & $1.0949\times$ & $1.0472\times$ $[1.0327,1.0604]$ & $9.0$ \\
$128$ & $-0.0082$ & $0.8207$ & $1.0973\times$ & $1.0468\times$ $[1.0328,1.0605]$ & $9.4$ \\
$256$ & $-0.0083$ & $0.8201$ & $1.0957\times$ & $1.0451\times$ $[1.0316,1.0583]$ & $9.9$ \\
$512$ & $-0.0093$ & $0.8158$ & $1.0981\times$ & $1.0488\times$ $[1.0345,1.0632]$ & $11.2$ \\
$1024$ & $-0.0078$ & $0.8164$ & $1.0996\times$ & $1.0519\times$ $[1.0380,1.0667]$ & $15.0$ \\
\bottomrule
\end{tabular}}
\end{table}

\subsection{Region-count ablation on 2DGS}
\label{ssec:exp_k_2dgs}
To test the region-count trend on a different representation, we repeat the
ablation on frozen 2DGS checkpoints for the same thirteen scenes, with
$\Kcl\in\{1,2,4,8,64,256,512,1024\}$ fixed before test evaluation.
\Cref{tab:k_ablation_2dgs} uses the exact $1/255$ opacity-bound path as its
baseline, isolating policy selection from the earlier geometric-bound
improvement. Under the same selection-view quality constraints, speedup rises
from $1.0838\times$ at $\Kcl=1$ to $1.1200\times$ at $\Kcl=8$, followed by
$1.1216\times$ and $1.1228\times$ at $\Kcl=64$ and $256$. The direct paired
$256$-versus-$64$ ratio is $1.0011\times$ with a scene-bootstrap interval of
$[0.9988,1.0037]$, while mean compilation time rises from $53.5$ to $57.9$ seconds.
The added $\Kcl=2$ and $4$ policies give $1.0144\times$ and
$1.0287\times$ over their same-batch $\Kcl=1$ controls. At the other end,
$\Kcl=512$ and $1024$ give direct paired speedups of $0.9999\times$
$[0.9956,1.0043]$ and $1.0002\times$ $[0.9938,1.0066]$ over the
same-batch $\Kcl=64$ control, with mean compilation times of $60.5$ and
$62.6$ seconds. Thus increasing the region count beyond $64$ has not
established an additional runtime benefit in this experiment; this does not
establish $\Kcl=64$ as universally optimal. For small $\Kcl$, the existing
construction uses $\min(8,\Kcl)$ risk strata, so the $2$- and $4$-group
policies also use fewer strata.
The policies attain different distortions under the common constraints, so
we report PSNR, SSIM, and LPIPS changes rather than claim exact quality matching.

\begin{table}[t]
\centering\small
\caption{\textbf{Region-count ablation on 2DGS, all thirteen scenes.}
All variants use the same frozen 30k-iteration 2DGS checkpoints at native
resolution, the same 16 calibration and 32 disjoint selection views, the
11-level ladder, $S_{\mathrm{eff}}=\min(8,\Kcl)$ risk strata, and 48 Lagrangian penalties with the same
proposal-generation procedure. Selection requires minimum-view teacher PSNR
$\ge48$\,dB and worst-view p99 $\le0.012$; these are common constraints,
not exact matching of attained distortion. The baseline is the 2DGS path
with the exact $1/255$ opacity bound, so the speedups exclude the prior
$3\sigma$-to-exact-bound transition. Quality deltas and pair ratio $R$ are
relative to this baseline on the full test views. Each measurement batch has five arms with cached frozen activations and
fifteen balanced same-process timing cycles. The added low-$\Kcl$ and
high-$\Kcl$ batches each include the same frozen $\Kcl=1$ and $64$
controls and the exact-bound baseline; each ratio uses its own batch's timing.
Speedups are scene geometric means; quality deltas, $R$, and compilation
times are arithmetic scene means. Brackets are 95\% scene-bootstrap intervals.
Compilation time excludes checkpoint loading.}
\label{tab:k_ablation_2dgs}
\setlength{\tabcolsep}{4pt}
\resizebox{\linewidth}{!}{%
\begin{tabular}{rrrrrrrr}
\toprule
$\Kcl$ & $\Delta$PSNR & $\Delta$SSIM & $\Delta$LPIPS$\downarrow$ & $R$ & vs. exact bound & vs. $\Kcl=1$ [95\% CI] & compile (s) \\
\midrule
$1$ & $-0.0049$ & $-1.21\!\times\!10^{-4}$ & $-7.07\!\times\!10^{-5}$ & $0.9161$ & $1.0838\times$ & $1.0000\times$ & $11.7$ \\
$2$ & $-0.0056$ & $-1.34\!\times\!10^{-4}$ & $-8.94\!\times\!10^{-5}$ & $0.9027$ & $1.0974\times$ & $1.0144\times$ $[1.0079,1.0218]$ & $18.0$ \\
$4$ & $-0.0068$ & $-1.61\!\times\!10^{-4}$ & $-1.00\!\times\!10^{-4}$ & $0.8882$ & $1.1129\times$ & $1.0287\times$ $[1.0169,1.0432]$ & $24.8$ \\
$8$ & $-0.0075$ & $-1.67\!\times\!10^{-4}$ & $-1.11\!\times\!10^{-4}$ & $0.8801$ & $1.1200\times$ & $1.0334\times$ $[1.0197,1.0485]$ & $37.7$ \\
$64$ & $-0.0093$ & $-1.83\!\times\!10^{-4}$ & $-1.06\!\times\!10^{-4}$ & $0.8787$ & $1.1216\times$ & $1.0349\times$ $[1.0215,1.0496]$ & $53.5$ \\
$256$ & $-0.0098$ & $-1.93\!\times\!10^{-4}$ & $-1.09\!\times\!10^{-4}$ & $0.8782$ & $1.1228\times$ & $1.0360\times$ $[1.0214,1.0520]$ & $57.9$ \\
$512$ & $-0.0092$ & $-1.70\!\times\!10^{-4}$ & $-1.11\!\times\!10^{-4}$ & $0.8797$ & $1.1186\times$ & $1.0355\times$ $[1.0232,1.0504]$ & $60.5$ \\
$1024$ & $-0.0104$ & $-1.90\!\times\!10^{-4}$ & $-1.06\!\times\!10^{-4}$ & $0.8786$ & $1.1188\times$ & $1.0357\times$ $[1.0220,1.0516]$ & $62.6$ \\
\bottomrule
\end{tabular}}
\end{table}

\subsection{Region-construction ablation}
\label{ssec:exp_partition}

The $\Kcl$ sweep establishes the value of heterogeneous control; we next hold
$\Kcl=64$ fixed and change only how Gaussians are assigned to decisions. Random
work-balanced groups and a robust $4\times4\times4$ uniform spatial grid test
whether the gain comes merely from adding 64 free parameters. To avoid
confounding construction with a different selected distortion, every family is
interpolated to the same 55\,dB teacher PSNR.
\Cref{tab:partition_ablation} separates risk stratification, Morton spatial
locality, and pair-workload balancing. Random and Uniform-grid retain
$88.3\%/88.2\%$ of pairs and reach only $1.065\times/1.066\times$; the full
risk-aware construction retains $80.8\%$ and reaches $1.103\times$. Full is
faster than Random by $1.036\times$ $[1.029,1.046]$ and than Uniform-grid by
$1.035\times$ $[1.029,1.041]$, winning all thirteen scenes in both comparisons.
Spatial, work-balanced regions without risk stratification lie between them at
$1.082\times$, and Full remains $1.020\times$ faster $[1.015,1.024]$ (13/13).
Thus the Pareto gain comes from grouping spatially coherent Gaussians by tail
efficiency, rather than merely exposing 64 independent decisions.

Workload balancing does what it is designed to do---it lowers region-workload
CV from $0.684$ to $0.047$---but does not add latency gain at matched quality:
count-balanced risk strata reach $1.1034\times$ versus the full method's
$1.1032\times$; their direct ratio is $0.9998\times$ with a confidence interval
that crosses one. We therefore treat workload balancing as
a conditioning and region-resolution regularizer, not as an independent source
of speedup.

\begin{table}[t]
\caption{\textbf{Region-construction ablation on all thirteen Speedy-Splat scenes.}
We fix $\Kcl=64$ decisions and keep probes, ladder, and search unchanged.
Each validation-selected Pareto curve is interpolated in teacher MSE to
55\,dB, with only its two bracketing policies timed in a common balanced
process. Random retains work balancing alone; Uniform grid retains spatial
locality alone; No-risk keeps Morton locality and work balancing; No-workload
keeps risk strata and Morton order but partitions by Gaussian count. Work CV
measures the coefficient of variation in baseline pair workload across regions.
Full/row above one indicates that Full is faster, and brackets give $95\%$
scene-bootstrap intervals.}
\label{tab:partition_ablation}
\centering\small
\setlength{\tabcolsep}{3pt}
\resizebox{\linewidth}{!}{%
\begin{tabular}{lcccrrrr}
\toprule
Grouping & Risk & Spatial & Work-bal. & Work CV & $R$ & vs. AccuTile & Full/row [95\% CI] (wins) \\
\midrule
Random & no & no & yes & $0.005$ & $0.8832$ & $1.0649\times$ & $1.0360\times$ $[1.0285,1.0458]$ (13/13) \\
Uniform grid & no & yes & no & $3.652$ & $0.8823$ & $1.0658\times$ & $1.0351\times$ $[1.0293,1.0408]$ (13/13) \\
No risk strata & no & yes & yes & $0.005$ & $0.8480$ & $1.0820\times$ & $1.0196\times$ $[1.0151,1.0243]$ (13/13) \\
No workload balance & yes & yes & no & $0.684$ & $0.8087$ & $1.1034\times$ & $0.9998\times$ $[0.9985,1.0011]$ (6/13) \\
Full & yes & yes & yes & $0.047$ & $0.8076$ & $1.1032\times$ & -- \\
\bottomrule
\end{tabular}}
\end{table}

\subsection{Why not simply raise the constant?}
\label{ssec:exp_global}

The most direct objection to this paper is that our gain merely reflects $\base=1/255$
being too conservative, and that setting it to $2/255$ or $4/255$ would capture it. We
answer that first, because it determines what the rest of the section is worth.

\Cref{tab:fixed} applies each fixed multiple of $\base$ to all thirteen scenes. Read by
the mean, the objection looks correct: at $4\base$ the average costs only $-0.085$\,dB
for $1.111\times$. Read by the worst scene it does not: the same $4\base$ costs
$-0.426$\,dB on \emph{bonsai}. And \emph{bonsai} is the least favourable scene at
\emph{every} multiple, so a single constant is not free to sit at its own average --- it
must be clamped to whichever scene happens to be hardest.

Making that clamp explicit answers the objection quantitatively. A fixed constant matched
to our worst-scene quality ($-0.022$\,dB) can only reach $1.62\base$ and $1.036\times$,
against our $1.1009\times$ in this dedicated sweep. A fixed constant matched to
that \emph{speed} must run at $3.6\base$, where the worst scene loses
$0.334$\,dB---$15\times$ our worst-case error.
Note also that our own worst scene is \emph{garden}, not \emph{bonsai}: the compiled
policy specifically relieves the scene that constrains the global rule.

There is a second, quieter point. Choosing $2\base$ or $4\base$ safely presupposes
knowing which scenes tolerate it, and the per-scene optimum found by our selector ranges
over $1.5$--$3.0\base$. Recovering even that scalar requires scene measurements
and rendered acceptance tests. We call the process compilation because it
automatically emits a frozen deployment asset; it replaces manual threshold
choice rather than claiming that no offline selection occurs.

\begin{table}[t]
\caption{``Why not simply raise the constant?'' Each row applies one fixed multiple of $\base$ to all thirteen scenes in one dedicated cutoff sweep. The mean is reassuring and the worst scene is not: the same scene (\emph{bonsai}) is the least favourable at every multiple, so a single constant must be clamped to it. Matching \textsc{Ours-Regional}'s worst-scene quality caps a fixed constant at $1.62\base$ and $1.036\times$; matching its $1.1009\times$ speed in this sweep forces the constant to $3.6\base$, where the worst scene loses $-0.334$\,dB against $-0.022$\,dB. The independent all-view twelve-cycle main retiming is $1.096\times$. Choosing safely requires measuring the scene---the compilation step already present in \textsc{Ours-Global}.}
\label{tab:fixed}
\centering\small
\begin{tabular}{lrrrr}
\toprule
Threshold & mean $\Delta$PSNR & worst $\Delta$PSNR & worst scene & vs.\ Official-Fixed \\
\midrule
$1.25\base$ & $-0.0011$ & $-0.0030$ & \emph{bonsai} & $1.0194\times$ \\
$1.5\base$ & $-0.0038$ & $-0.0126$ & \emph{bonsai} & $1.0305\times$ \\
$2\base$ & $-0.0126$ & $-0.0493$ & \emph{bonsai} & $1.0530\times$ \\
$3\base$ & $-0.0425$ & $-0.2005$ & \emph{bonsai} & $1.0855\times$ \\
$4\base$ & $-0.0853$ & $-0.4255$ & \emph{bonsai} & $1.1113\times$ \\
$6\base$ & $-0.1900$ & $-0.9404$ & \emph{bonsai} & $1.1497\times$ \\
$8\base$ & $-0.2889$ & $-1.3708$ & \emph{bonsai} & $1.1767\times$ \\
\midrule
\textbf{Ours-Regional} & $\mathbf{-0.0074}$ & $\mathbf{-0.0218}$ & \emph{garden} & $\mathbf{1.1009\times}$ \\
\bottomrule
\end{tabular}
\end{table}

\paragraph{Calibration-view sensitivity.} Region count is swept on all thirteen
scenes in \Cref{tab:k_ablation}; \Cref{tab:calibration_views} likewise varies
calibration views over $4/8/16/32$ while matching every curve at 55\,dB. All
four settings have aggregate intervals above one versus AccuTile, while every
direct interval against the 16-view default crosses one. Increasing from 16 to
32 views lowers $R$ from $0.808$ to $0.800$ but yields statistically
indistinguishable latency, so we retain 16 rather than tune this value per
scene. Across three independent random 16/32 calibration/selection splits,
speedup is $1.0945\times\pm0.0010\times$, pair ratio is $0.8299\pm0.0023$, and
mean PSNR change is $-0.0085\pm0.0012$\,dB
(\Cref{tab:split_stability}).

\begin{table}[t]
\caption{\textbf{Calibration-view sensitivity on all thirteen Speedy-Splat scenes.} Each family uses $\Kcl=64$ and 32 disjoint selection views; only the number of calibration views changes. Curves are interpolated to 55\,dB teacher PSNR and their bracketing policies are timed together. Brackets bootstrap scenes and parentheses count wins.}
\label{tab:calibration_views}
\centering\small
\setlength{\tabcolsep}{3pt}
\resizebox{\linewidth}{!}{%
\begin{tabular}{rrrrrl}
\toprule
Calibration views & $\Delta$PSNR & $R$ & vs. AccuTile [95\% CI] (wins) & $16$/row [95\% CI] (wins) & Compile (s) \\
\midrule
$4$ & $-0.0095$ & $0.8661$ & $1.0827\times$ $[1.0579,1.1079]$ (13/13) & $1.0036\times$ $[0.9781,1.0257]$ (9/13) & $12.0$ \\
$8$ & $-0.0109$ & $0.8342$ & $1.0774\times$ $[1.0260,1.1197]$ (11/13) & $1.0085\times$ $[0.9831,1.0358]$ (7/13) & $15.4$ \\
$\mathbf{16}$ & $-0.0099$ & $0.8076$ & $1.0866\times$ $[1.0542,1.1160]$ (11/13) & -- & $9.0$ \\
$32$ & $-0.0074$ & $0.8000$ & $1.0889\times$ $[1.0523,1.1238]$ (11/13) & $0.9979\times$ $[0.9792,1.0229]$ (3/13) & $17.2$ \\
\bottomrule
\end{tabular}}
\end{table}

\subsection{Dense, flow-aligned temporal evaluation}
\label{ssec:exp_temporal}

We construct a continuous path by inserting three SE(3)-interpolated cameras
between every adjacent pair of official test poses. This yields 359 frames over
\emph{bicycle}, \emph{bonsai}, and \emph{counter}. For each method, let the
pruning residual be its rendered image minus the unmodified AccuTile image. We
estimate bidirectional RAFT-Small flow~\citep{teed2020raft} between consecutive
AccuTile frames, backward-warp the preceding residual, reject occlusions by
forward--backward consistency, and measure the remaining residual change. The
\textsc{Ours-Global} and \textsc{Ours-Regional} policies are chosen closest to 55\,dB using held-out training
views; neither test images nor the interpolated path participate in selection.

Region adaptation lowers the geometric-mean aligned residual change to
$0.789\times$ the \textsc{Ours-Global} value while also reducing the pair ratio from
$0.873$ to $0.775$. Because lower per-frame error alone can lower this temporal
quantity, we also normalize each policy's aligned change by its own frame RMSE.
That ratio is $0.990\times$ overall: no aggregate temporal regression, but no
claim of an improvement either. The negative \emph{counter} result
($1.050\times$ normalized) is retained in \Cref{tab:temporal_flow}; the evidence
supports aggregate stability on these paths, not a universal no-popping guarantee.

\begin{table}[t]
\centering
\caption{\textbf{Dense, flow-aligned temporal evaluation.}
Before constructing test-camera paths, we select \textsc{Ours-Global} and
\textsc{Ours-Regional} near 55\,dB on held-out training views. We insert three SE(3)-interpolated views between adjacent test poses, masking occlusions
with bidirectional RAFT-Small consistency. The flow-aligned residual-change
RMSE is $E$ ($\times 10^{-3}$); $E_n$ normalizes it by each policy's frame
RMSE. Lower $E$ and Regional/Global are better. Aggregates are
geometric means, with valid flow weighted by transition.}
\label{tab:temporal_flow}
\small
\resizebox{\linewidth}{!}{%
\begin{tabular}{lccccccc}
\toprule
Scene & Frames & Valid flow & $R$ Global & $R$ Regional & $E$ Global & $E$ Regional & $E_n$ Regional/Global \\
\midrule
Bicycle & 97 & 45.8\% & 0.922 & 0.815 & 2.700 & 1.798 & 0.926$\times$ \\
Bonsai & 145 & 77.4\% & 0.898 & 0.772 & 2.081 & 1.373 & 0.998$\times$ \\
Counter & 117 & 79.6\% & 0.803 & 0.738 & 1.386 & 1.549 & 1.050$\times$ \\
\midrule
Aggregate & 359 & 69.6\% & 0.873 & 0.775 & 1.982 & 1.564 & 0.990$\times$ \\ 
\bottomrule
\end{tabular}}
\end{table}

\subsection{Transfer to independently optimized rasterizers}
\label{ssec:exp_transfer}

The strongest portability test is to run the Speedy-Splat policies underlying \Cref{tab:main}
inside software we did not write, without recalibration. We modify only the
threshold lookup in the official SeeLe~\citep{huang2026seele} and
FlashGS~\citep{feng2025flashgs} rasterizers. SeeLe uses an axis-aligned footprint
and a different blending path; FlashGS combines precise Gaussian intersection,
adaptive scheduling, and pipelining.

\Cref{tab:transfer} reports both transfers under the same twelve-cycle protocol
as the main table. SeeLe becomes $1.091\times$ faster at standard dataset resolution and
$1.260\times$ at 4K. FlashGS improves by $1.121\times$ and $1.317\times$, and
the regional policy beats transferred \textsc{Ours-Global} by $1.061\times$ and
$1.138\times$. The FlashGS result is especially useful because it
shows the policy still removes work after a strong geometric tile-intersection
optimization. These are transfer results, not claims against the full SeeLe
camera-clustering pipeline or FlashGS's published end-to-end benchmark.

\begin{table}[t]
\caption{\textsc{Ours-Global} and \textsc{Ours-Regional} policies compiled on AccuTile and transferred unchanged into two official optimized rasterizers. FlashGS retains its precise intersection and scheduling optimizations. $R$ and teacher PSNR are measured against each rasterizer's native path. Every row uses twelve balanced same-process timing cycles; no row recalibrates or fine-tunes either policy.}
\label{tab:transfer}
\centering\small
\begin{tabular}{llrrrrr}
\toprule
Rasterizer & Resolution & Scenes & $R$ & Teacher PSNR & vs.\ Official-Fixed & vs.\ Ours-Global \\
\midrule
SeeLe & Standard & 13 & 0.8003 & 54.75 & $\mathbf{1.0908\times}$ & $1.0437\times$ \\
SeeLe & 4K ($3840$\,px) & 13 & 0.7539 & 56.41 & $\mathbf{1.2602\times}$ & $1.1233\times$ \\
\midrule
FlashGS & Standard & 13 & 0.8205 & 52.52 & $\mathbf{1.1214\times}$ & $1.0605\times$ \\
FlashGS & 4K ($3840$\,px) & 13 & 0.7723 & 50.38 & $\mathbf{1.3166\times}$ & $1.1379\times$ \\
\bottomrule
\end{tabular}
\end{table}

\subsection{The cost model predicts, and one of its predictions is negative}
\label{ssec:exp_cost}

\begin{figure}[t]
  \centering
  \includegraphics[width=\linewidth]{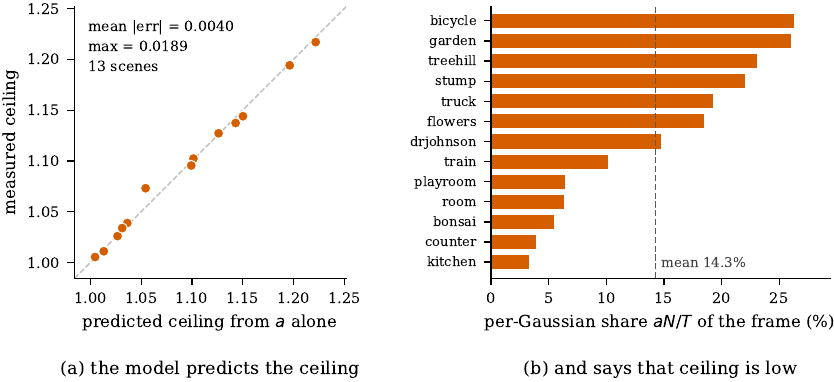}
  \caption{\textbf{A prediction made before the implementation, and then tested.} \emph{(a)}
  Fitting only $a$ --- from a single oracle-subset comparison taking seconds --- predicts the
  measured ceiling of region pre-culling to a mean absolute error of $0.0040$ over thirteen
  scenes. \emph{(b)} The prediction is unfavourable: the per-Gaussian term is $14.3\%$ of the
  honest frame on average, so the tested perfect, free oracle is capped near
  $1.17\times$. The estimate is specific to this checkpoint family, resolution, GPU, and
  pre-culling design.}
  \label{fig:costmodel}
\end{figure}

\Cref{sec:costmodel} fits an empirical affine model within controlled sweeps;
here we test whether it is useful for dispatch. An oracle visible-subset
comparison gives an estimated upper bound for the tested region pre-culler
before implementing its runtime path. Across thirteen scenes, the estimate
matches the realized ceiling to a mean absolute error of $0.004$
(\Cref{fig:costmodel}a). The result is unfavourable: the per-Gaussian term is
only $14.3\%$ of the honest frame on average ($3.3$--$26.3\%$), so the tested
perfect, free oracle is capped near $1.17\times$ and the implemented culler---
conservative predicate and planning included---returns $1.010\times$ overall. We report this as a negative result rather than omitting
it, because it is the same model that produces the positive one.

The model also says \emph{where} the negative result reverses. Dense stress-test
checkpoints have a larger measured $a\Npts/T$ share, even though their pair count
also changes. \Cref{tab:stack} and \Cref{fig:dispatch}b show that on the tested
checkpoints above ${\sim}2$M Gaussians the two mechanisms compose nearly
multiplicatively, reaching $1.235\times$ at standard dataset resolution, and the
pre-culling arm is image-bit-exact in all $52$ runs. Two rows use legacy packaged
checkpoints and are deliberately not attributed to named model families. Below
${\sim}500$K Gaussians pre-culling costs more than it saves
($0.988\times$ on Speedy-Splat) and the dispatch rule of \Cref{sec:dispatch}
switches it off. At 4K the measured per-Gaussian share falls to $3$--$12\%$ of
the frame, so the planning-cost test selects the policy alone.

\begin{table}[t]
\caption{Policy truncation primarily reduces pair-correlated work; image-bit-exact region pre-culling skips conservatively rejected Gaussian blocks before projection. Standard dataset resolution, planning cost included. This five-arm stress microbenchmark uses at most twelve evenly spaced test views per scene, so its policy column is not a replacement for the all-view twelve-cycle main-table timing. The pre-culling arm's maximum absolute pixel difference against the unmodified render is $0$ in all $52$ runs, so the quality cost of the combination is that of the policy alone. Legacy dense A/B are historical packaged checkpoints used only as density stress tests, not as evidence for a named model family. Below roughly $500$K Gaussians the frustum test costs more than it saves and the dispatch rule of \Cref{sec:dispatch} disables it.}
\label{tab:stack}
\centering\small
\begin{tabular}{lrrrrrr}
\toprule
Checkpoint & Mean Gaussians & Base (ms) & Policy & Pre-cull & \textbf{Both} & Gain from stacking \\
\midrule
Legacy dense A & 2,933,368 & 3.63 & $1.080\times$ & $1.056\times$ & $\mathbf{1.142\times}$ & $1.058\times$ \\
Legacy dense B & 2,519,315 & 2.52 & $1.138\times$ & $1.080\times$ & $\mathbf{1.235\times}$ & $1.085\times$ \\
DashGaussian & 2,167,600 & 2.47 & $1.144\times$ & $1.063\times$ & $\mathbf{1.220\times}$ & $1.067\times$ \\
Speedy-Splat & 407,593 & 0.78 & $1.132\times$ & $0.988\times$ & $\mathbf{1.115\times}$ & $0.984\times$ \\
\bottomrule
\end{tabular}
\end{table}

\subsection{Cost}
\label{ssec:exp_cost_detail}
Compilation reads the checkpoint and calibration views, runs both probes, and solves selection in $9.0$\,s per scene on average for the default $\Kcl=64$ (at most $14.0$\,s, excluding checkpoint load). The sensitivity probe invokes backward-mode differentiation but updates no checkpoint parameter and needs no optimizer state. The exported policy is one \texttt{uint8} level per Gaussian plus a table of at most $256$ thresholds: exactly $\Npts$ bytes for the indices and under $1$\,KB for the table. The compact Speedy-Splat checkpoints occupy $0.14$--$0.88$\,MB; dense ports scale linearly beyond that range. Inference reads the byte during tile enumeration, with no extra kernel or per-frame policy inference. The cost amortizes after roughly $16$k rendered frames at 4K and $126$k at standard dataset resolution.

\subsection{Limitations} \label{ssec:exp_limitations}

\emph{Calibration needs the training views.} Compilation measures pair savings and
tail energy by rendering, so it needs the camera poses and images the checkpoint was
fitted to. It is a per-scene calibration step, not a data-free transform of an
arbitrary released \texttt{.ply}. It does not need the optimizer state or the training
run, and it performs no parameter update.

\emph{Scope is threshold-based tile enumeration.} The method assumes a rasterizer
whose footprint is governed by a contribution threshold. It applies directly to
AccuTile-style rasterizers, to SeeLe, and to 2DGS, Scaffold-GS, or vanilla 3DGS
once their $3\sigma$ box is replaced by such a bound. Oriented-box, convex-hull and non-tiled rasterizers expose a
different control surface and are out of scope.

\emph{Regions do not beat primitives.} As \Cref{tab:matched} shows, $\Kcl$
region-shared thresholds tie per-Gaussian control on speed. They reduce the
number of independent search decisions and may regularize selection, but our
current packed format stores one byte per Gaussian for either policy. We make no
compression or superiority claim for regional control.

\emph{Static policies and camera motion.} The exported policy is fixed. Our
dense flow-aligned test covers 359 frames on three scenes and has no aggregate
regression, but \emph{counter} regresses by $5.0\%$ after normalization. Broader
trajectories and a perceptual user study would be needed for a universal
no-popping claim.

\emph{Held-out constraints are not test guarantees.} Three random
calibration/selection splits produce nearly identical aggregate speed and pair
ratios, but the per-Gaussian multiplier vectors agree exactly only $49.5\%$ of
the time. On unseen test views, 18 of the 39 split--scene runs exceed the
$0.012$ selection cap, eight reach at least $0.015$, and the maximum is $0.032$.
The cap controls held-out training views; it does not certify every unseen pose.
Selection and test diagnostics both use floating-point residuals and the same
$4\times$ spatial subsample, so the excesses are not a dtype mismatch.

\emph{Tile-aligned residual structure.} Lossy support truncation removes a
Gaussian from a whole tile at once. In a three-scene, 92-test-view diagnostic,
the residual finite-difference at the true 16-pixel tile phase is the largest
of all 16 phases in every view. Its mean amplitude is only $0.19$--$0.31$ of one
8-bit code value, but its mean boundary/non-boundary ratio is
$5.86$--$6.18\times$ (\Cref{app:tile_boundary}). This establishes structured
error, not perceptual visibility; a controlled user study remains absent.

\emph{One GPU class.} All timings are from RTX~5000 Ada GPUs; the four locally
available devices are the same model. The cost model suggests the balance between
the two mechanisms shifts on hardware with a different memory-to-compute ratio,
but a second desktop or mobile hardware class remains unmeasured.

                                \section{Backbone-transition attribution}
\label{app:backbone_decomposition}

The main table intentionally presents the deployment-facing comparison
``official model versus model + Ours.'' \Cref{tab:backbone_decomposition} provides
the accounting needed to interpret that compact presentation. Speedy-Splat
already exposes the exact $1/255$ footprint, so its full gain is due to our
compiler. The public 2DGS and Scaffold-GS rasterizers start from $3\sigma$;
installing an exact opacity-aware bound contributes the first speedup stage and
is prior work, while $\Kcl=1$ and $\Kcl=64$ are our compiler stages.

\begin{table}[t]
\caption{\textbf{Attribution behind the simplified backbone comparison in \Cref{tab:main}.} Exact-$1/255$ replaces a released $3\sigma$ tile box by the opacity-aware bound already established by prior work~\citep{radl2024stopthepop,hanson2025speedysplat}; we do not count that transition as a compiler contribution. $\Kcl=1$ and $\Kcl=64$ are our compiler with one and 64 decisions. Values are speedups over the official released path except for the final incremental column.}
\label{tab:backbone_decomposition}
\centering\small
\setlength{\tabcolsep}{5pt}
\begin{tabular}{lrrrr}
\toprule
Backbone & Exact-$1/255$ & Ours ($\Kcl=1$) & Ours ($\Kcl=64$) & $\Kcl{=}64/\Kcl{=}1$ \\
\midrule
Speedy-Splat & $1.000\times$ & $1.048\times$ & $1.095\times$ & $1.047\times$ \\
2DGS & $1.313\times$ & $1.422\times$ & $1.474\times$ & $1.037\times$ \\
Scaffold-GS & $1.252\times$ & $1.301\times$ & $1.320\times$ & $1.014\times$ \\
\bottomrule
\end{tabular}
\end{table}

\subsection{Threshold semantics and low-opacity Gaussians}
\label{app:threshold_semantics}
The exported lookup table stores absolute thresholds $m_g\base$. The default
Speedy-Splat path uses these thresholds directly, without an opacity-relative
cap; a threshold at or above Gaussian opacity yields no tile pairs. The 2DGS
and Scaffold-GS threshold-bound ports likewise reject primitives whose opacity
cannot exceed the selected threshold. This changes their participation in a
render, not the stored checkpoint parameters. The FastGS port instead applies
$\min(m_g\base,0.95\alpha_g)$ and subtracts $10^{-6}$ when the cap binds before
constructing its backend-specific extent. Thus preserving nonempty support is
not a method-wide invariant, and an all-one policy should not be assumed to
reproduce every backend's native pair list merely from its multiplier values.

A synthetic check of the default Speedy-Splat path used five opacities spanning
$0.5\base$ to $0.5$. Both scalar and packed all-one thresholds reproduced the
baseline images bit-for-bit and preserved pair counts in these cases. For an
opacity of $0.5\base$, the baseline and both all-one paths emitted zero pairs;
the capped FastGS scalar and packed all-one paths emitted four pairs versus
zero for their native baseline, while pixel-level alpha rejection kept all
three images identical. This distinguishes image equivalence from work equivalence;
it is a bounded regression check, not a universal floating-point proof.

\section{The checkpoint families used by Fragment Pruning}
\label{app:fragment_backbones}

Fragment Pruning~\citep{ye2024fragmentpruning} reports its learned per-Gaussian
fragment cutoff on vanilla 3DGS and Mini-Splatting checkpoints. We therefore
apply TileSkipper independently to clean official-style 30k 3DGS checkpoints
and official Mini-Splatting outputs over the same thirteen scenes. This is a
checkpoint-composition test, not an implementation of Fragment Pruning and not
a cross-hardware speed contest: their published FPS uses a Jetson Orin NX and
an OpenGL renderer, whereas \Cref{tab:fragment_backbones} uses our CUDA
rasterizer on RTX~5000 Ada.

At standard dataset resolution, the deployment-facing totals are
$1.493\times$ for 3DGS and $1.346\times$ for Mini-Splatting; at fixed 1080p
they rise to $1.841\times$ and $1.599\times$, with all thirteen scenes faster
in both settings. Those totals must not be attributed entirely to our compiler:
at standard resolution, installing the prior exact $1/255$ bound already gives
$1.454\times$ and $1.266\times$, respectively. The regional policy contributes
a further $1.027\times$ and $1.063\times$ ($1.038\times$ and $1.089\times$ at
1080p). Thus the
experiment supports compatibility with dense and aggressively pruned
checkpoints, but it does not establish superiority to Fragment Pruning's
learned cutoff. Their reported dataset-level ratios are in a similar numerical
range, but the hardware, renderer, and 60-FPS cap make direct ranking invalid.

Every selected policy passes the $48$-dB and $0.012$ p99 constraints on the
disjoint held-out training views used by the compiler. On unseen test views,
mean ground-truth PSNR changes remain only $-0.010/-0.014$ dB for
3DGS/Mini-Splatting, but at fixed 1080p Mini-Splatting's worst test view reaches
$45.76$ dB teacher PSNR and $0.0259$ p99. This is a concrete instance of the
held-out-versus-test limitation in \Cref{ssec:exp_limitations}; we report it
rather than treating the selection constraint as a certificate.

\begin{table}[t]
\caption{\textbf{TileSkipper on the two checkpoint families used by Fragment Pruning.}
Standard-resolution and fixed $1920{\times}1080$ rendering on RTX~5000 Ada; each row uses the same patched
vanilla-3DGS CUDA rasterizer and twelve balanced timing cycles. ``Base $\rightarrow$
+Ours'' is the deployment-facing total; ``AccuTile $\rightarrow$ +Ours'' isolates
our compiled policy after the prior $3\sigma\rightarrow1/255$ exact-bound transition.
Intervals bootstrap scenes and parentheses count wins. These are not head-to-head
Jetson/OpenGL measurements against Fragment Pruning.}
\label{tab:fragment_backbones}
\centering\scriptsize
\setlength{\tabcolsep}{2.8pt}
\resizebox{\linewidth}{!}{%
\begin{tabular}{lllrrrr}
\toprule
Checkpoint & Resolution & Dataset & Base $\rightarrow$ +Ours [95\% CI] (wins) & AccuTile $\rightarrow$ +Ours & $3\sigma\rightarrow$AccuTile & $\Delta$PSNR \\
\midrule
3DGS & standard & All 13 & $\mathbf{1.493\times}$ $[1.350,1.656]$ (13/13) & $\mathbf{1.027\times}$ $[1.014,1.042]$ (13/13) & $1.454\times$ & $-0.003$ \\
     & 1080p & Mip-NeRF 360 & $1.729\times$ $[1.532,1.959]$ (9/9) & $1.042\times$ $[1.023,1.064]$ (9/9) & $1.659\times$ & $-0.014$ \\
     & & Tanks\&Temples & $2.246\times$ $[2.013,2.506]$ (2/2) & $1.027\times$ $[1.020,1.035]$ (2/2) & $2.186\times$ & $-0.009$ \\
     & & Deep Blending & $2.003\times$ $[1.947,2.060]$ (2/2) & $1.028\times$ $[1.021,1.036]$ (2/2) & $1.948\times$ & $+0.003$ \\
     & & All 13 & $\mathbf{1.841\times}$ $[1.657,2.035]$ (13/13) & $\mathbf{1.038\times}$ $[1.024,1.054]$ (13/13) & $1.774\times$ & $-0.010$ \\
\midrule
Mini-Splatting & standard & All 13 & $\mathbf{1.346\times}$ $[1.260,1.451]$ (13/13) & $\mathbf{1.063\times}$ $[1.046,1.081]$ (13/13) & $1.266\times$ & $-0.008$ \\
     & 1080p & Mip-NeRF 360 & $1.593\times$ $[1.477,1.733]$ (9/9) & $1.090\times$ $[1.069,1.115]$ (9/9) & $1.462\times$ & $-0.017$ \\
     & & Tanks\&Temples & $1.770\times$ $[1.748,1.792]$ (2/2) & $1.081\times$ $[1.071,1.091]$ (2/2) & $1.637\times$ & $-0.013$ \\
     & & Deep Blending & $1.473\times$ $[1.410,1.538]$ (2/2) & $1.095\times$ $[1.077,1.114]$ (2/2) & $1.345\times$ & $-0.004$ \\
     & & All 13 & $\mathbf{1.599\times}$ $[1.505,1.703]$ (13/13) & $\mathbf{1.089\times}$ $[1.074,1.107]$ (13/13) & $1.468\times$ & $-0.014$ \\
\bottomrule
\end{tabular}}
\end{table}

                                \section{Calibration-view split stability}
\label{app:split_stability}

To test whether the compiler depends on one favorable view partition, we repeat
the full $\Kcl=64$ Speedy-Splat experiment with random seeds 17, 43, and 89. Each
seed independently samples 16 calibration and 32 disjoint selection views in
every scene; the ladder, probes, constraints, and search are unchanged. The
three policies are then evaluated and timed with AccuTile in the same four-arm
process. \Cref{tab:split_stability} shows only $0.0010\times$ standard deviation
in the thirteen-scene speedup, and all three splits improve all thirteen scenes.
The exact per-Gaussian multiplier agreement across split pairs is $49.5\%$:
different view samples find different near-equivalent policies, while their
aggregate distortion, work, and latency remain stable. Mean compilation time is
$9.1$ seconds per scene. This aggregate stability does not extend to the
test-view tail constraint: 18 of 39 split--scene runs exceed p99 $0.012$, eight
reach at least $0.015$, and the maximum is $0.032$.

\begin{table}[t]
\caption{\textbf{Calibration/selection split stability on all thirteen Speedy-Splat scenes.} Each row independently samples 16 calibration and 32 disjoint selection views; $\Kcl=64$ and all compiler settings are fixed. Speedup intervals bootstrap scenes.}
\label{tab:split_stability}
\centering\small
\setlength{\tabcolsep}{5pt}
\resizebox{\linewidth}{!}{%
\begin{tabular}{lrrrrr}
\toprule
Split seed & $\Delta$PSNR & $R$ & vs. AccuTile & Wins & Compile (s) \\
\midrule
17 & $-0.0097$ & $0.8325$ & $1.0934\times$ $[1.0709,1.1176]$ & $13/13$ & $9.1$ \\
43 & $-0.0073$ & $0.8283$ & $1.0947\times$ $[1.0752,1.1160]$ & $13/13$ & $9.1$ \\
89 & $-0.0084$ & $0.8288$ & $1.0953\times$ $[1.0762,1.1178]$ & $13/13$ & $9.1$ \\
\bottomrule
\end{tabular}}
\end{table}

\paragraph{A scalar safety margin helps but does not solve view transfer.}
We reselect from each split's already-rendered candidate table under validation
p99 caps of $0.010$ and $0.008$, without consulting test views. As
\Cref{tab:p99_margin} shows, the tighter caps reduce violations from $18$ to
$5$ and $3$ of 39 runs, respectively, while increasing the retained-pair ratio
from $0.830$ to $0.877$. All three failures at $0.008$ are \texttt{kitchen};
its worst p99 remains $0.021$--$0.030$ across the three splits. Thus a global
margin trades work for robustness but does not address a view-coverage failure.

\begin{table}[t]
\caption{\textbf{Validation-cap guard-band diagnostic over the same 39 randomized
split--scene candidate tables.} We reselect policies using held-out training-view
measurements alone, then evaluate them on official test views. The original
$0.012$ row reproduces the reported policies. Tightening the validation cap
reduces unseen-view violations at the cost of retaining more pairs, while
\texttt{kitchen} failures persist. This post-hoc diagnostic is not used to select
the paper's primary operating point.}
\label{tab:p99_margin}
\centering\small
\setlength{\tabcolsep}{4pt}
\resizebox{\linewidth}{!}{%
\begin{tabular}{lrrrrr}
\toprule
Validation p99 cap & Test p99 $>0.012$ & Test p99 $\ge0.015$ & Worst test p99 & Mean $R$ & Mean worst-view teacher PSNR \\
\midrule
$0.012$ & $18/39$ & $8/39$ & $0.0320$ & $0.830$ & $52.06$\,dB \\
$0.010$ & $5/39$  & $5/39$ & $0.0305$ & $0.851$ & $54.08$\,dB \\
$0.008$ & $3/39$  & $3/39$ & $0.0301$ & $0.877$ & $56.43$\,dB \\
\bottomrule
\end{tabular}}
\end{table}

                                \section{Spatial heterogeneity of safe global thresholds}
\label{app:local_tolerance}

\begin{figure}[h]
  \centering
  \includegraphics[width=\linewidth]{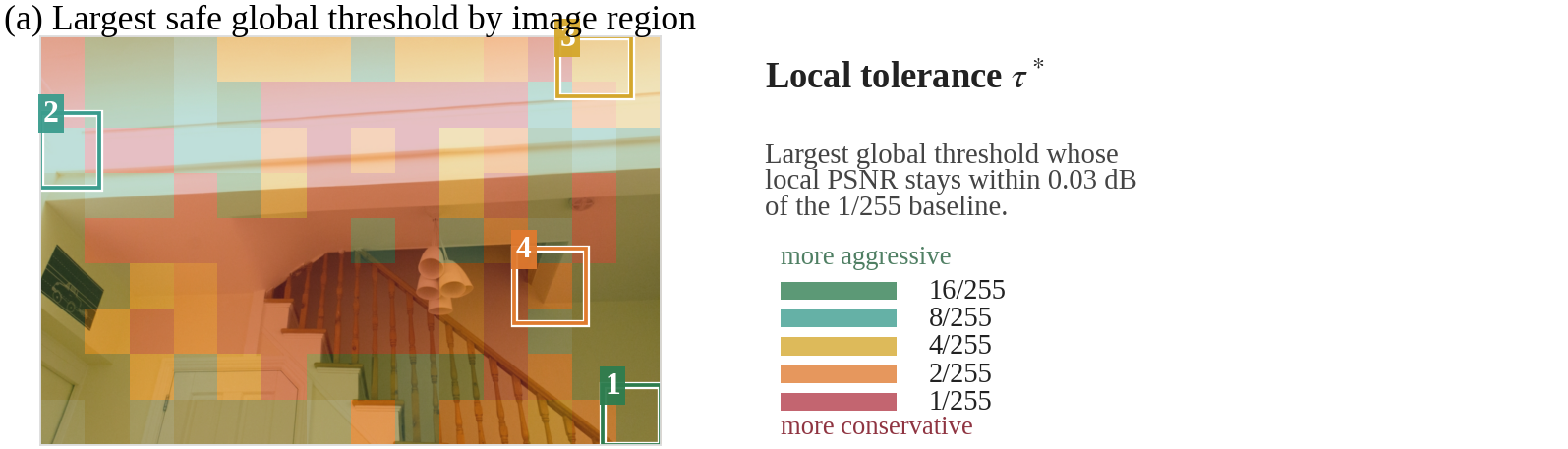}
  \caption{\textbf{Largest safe global tile threshold varies by
  $16\times$ across regions of a single view.} Each grid cell of a
  representative \emph{playroom} render is colored by the largest
  $\thr^\star\in\{1,2,4,8,16\}/255$ whose local PSNR stays within
  $0.03$\,dB of the $1/255$ baseline. The four highlighted examples are
  at \textbf{(1)} $16/255$ (low detail),
  \textbf{(2)} $8/255$ (material patch), \textbf{(3)} $4/255$ (fine detail),
  and \textbf{(4)} $2/255$ (edge or boundary). Other grid cells include the
  conservative $1/255$ value, giving the reported full spread. This is a local
  tolerance diagnostic rather than an evaluation of the compiled partition.}
  \label{fig:local_tolerance}
\end{figure}

\Cref{fig:local_tolerance} provides the pixel-level evidence behind
the Introduction's claim that a single global threshold is forced
into a compromise: it is too loose on easy regions and too tight on
hard ones. We partition a representative \emph{playroom} view into
a coarse spatial grid; for each region we sweep a global threshold
$\thr\in\{1,2,4,8,16\}/255$ and report the largest
$\thr^\star$ whose local PSNR (computed over the region's pixels
only) stays within $0.03$\,dB of the $1/255$ baseline. Each region
is colored by its $\thr^\star$. The grid contains the full $16\times$ spread
from $1/255$ to $16/255$ on a \emph{single view}. The four callouts are selected
at distinct values: region~1 is a low-detail example at $16/255$, region~2 a
material patch at $8/255$, region~3 a fine-detail example at $4/255$, and
region~4 an edge or boundary at $2/255$. Cells at $1/255$ appear in the map but
are not among these four callouts. This diagnostic establishes local
heterogeneity; it does not by itself measure how much of that heterogeneity a
particular compiled partition captures. Indeed, \emph{playroom} is one of two
scenes whose selected Regional policy collapses to a single level; pixel-local
heterogeneity is therefore not sufficient evidence of deployable regional gain.

\section{Tile-boundary residual structure}
\label{app:tile_boundary}

Tile support is a discrete action: when a tail no longer intersects a tile, all
of its fragments in that tile disappear together. We therefore test for the
specific grid artifact that aggregate PSNR, SSIM, and LPIPS may hide. On
\texttt{bicycle}, \texttt{bonsai}, and \texttt{counter}, we choose real Global,
Regional, and per-Gaussian policies nearest 55\,dB using validation metadata
only, then render all 92 official test views. For each candidate-minus-backbone
residual, we measure the RGB finite-difference magnitude at every pixel phase
modulo 16.

\begin{figure}[h]
  \centering
  \includegraphics[width=\linewidth]{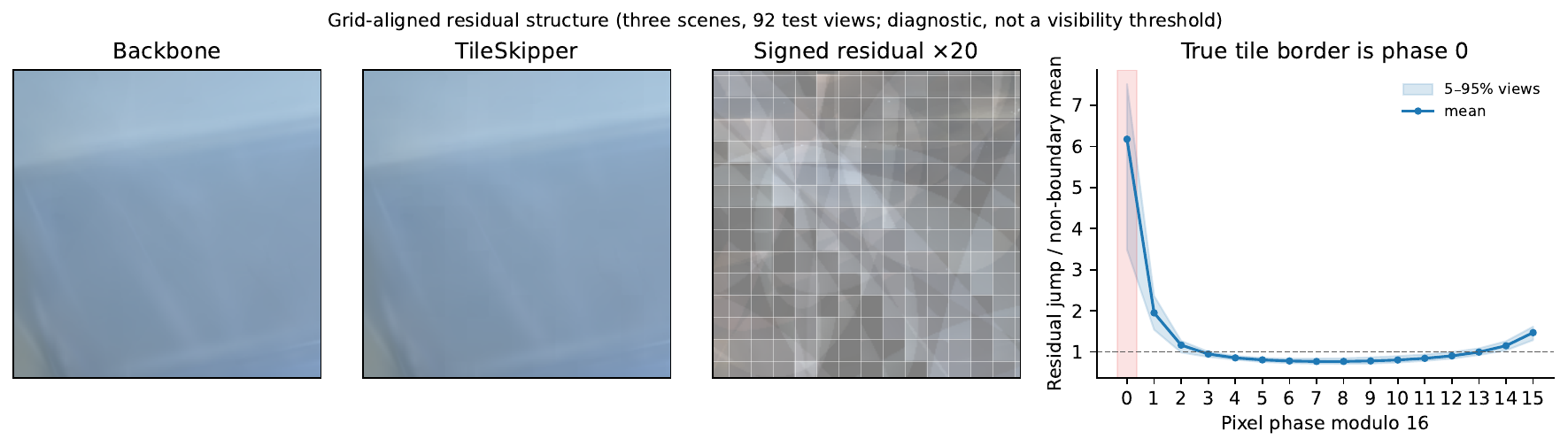}
  \caption{\textbf{Grid-aligned residual diagnostic.} The crop is selected by
  maximum boundary-weighted residual energy and the signed residual is amplified
  $20\times$; white lines show the true tile grid. The phase plot pools the
  Regional policy over three scenes and 92 test views. The highlighted phase zero
  is the actual tile border. The plot diagnoses residual structure and is not a
  perceptual visibility threshold.}
  \label{fig:tile_boundary}
\end{figure}

As shown in \Cref{fig:tile_boundary}, the true tile boundary (phase zero) is the largest of the 16 phases in all 92
views for every policy. The pooled mean boundary/non-boundary ratios are
$5.86\times$, $6.18\times$, and $5.93\times$ for Global, Regional, and
per-Gaussian control, respectively. Their absolute mean boundary jumps are
$0.192$, $0.287$, and $0.312$ of one 8-bit code value. Thus lossy tile-support
truncation does create grid-aligned residual structure, but this diagnostic
alone does not establish that it is visible at normal magnification. The
similar result for all three controls also shows that the effect is tied to the
tile-level action rather than regional grouping specifically.

\section{Qualitative comparison}
\label{app:qualitative}

\begin{figure}[h]
  \centering
  \includegraphics[width=\linewidth]{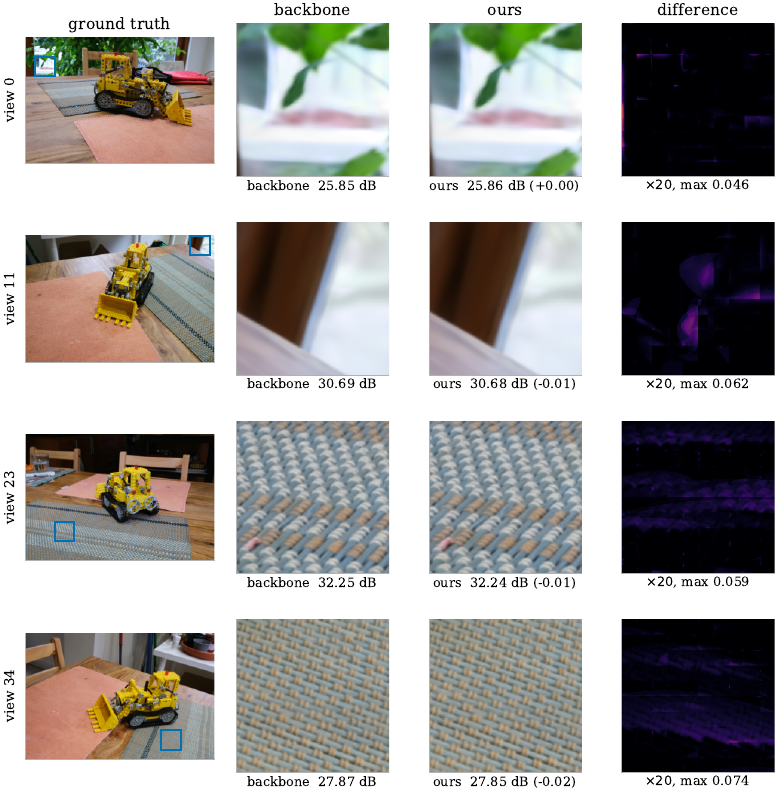}
  \caption{\textbf{Backbone versus the compiled policy on held-out \texttt{kitchen} views.}
  Crops are placed automatically at the window of largest disagreement, not chosen by hand;
  the difference column is amplified $20\times$. PSNR is against ground truth, so it is
  comparable to \Cref{tab:main}.}
  \label{fig:qualitative}
\end{figure}

\Cref{fig:qualitative} shows the backbone and the compiled policy on four held-out
\texttt{kitchen} test views. For each view we crop the $160\times160$ window where the two
renders differ \emph{most}, so the figure shows the hardest region rather than a
representative one, and amplify the difference by $20\times$ to make it visible at all.
Per-view PSNR against ground truth changes by $+0.00$ to $-0.02$\,dB and the largest
single-pixel difference is $0.074$ on a $[0,1]$ scale---roughly $19$ quantization steps
of an $8$-bit channel, concentrated on the woven placemat and the specular edge of the
window frame. Their concentration in high-frequency content shows where even a
conservative allocation remains vulnerable; it should not be read as evidence that the
compiler intentionally assigns a looser cutoff to those regions.

\newpage

\end{document}